\documentclass[journal=pasa, manuscript=r&d-internship-report, year=2022, volume=37]{cup-journal}

\usepackage{microtype,siunitx,booktabs}
\usepackage[utf8]{inputenc}
\usepackage{longtable} 
\usepackage{amsfonts} 
\usepackage{amsmath}
\usepackage[draft]{minted} 
\usepackage[style=numeric, sorting=none]{biblatex} 
\newcommand{\norm}[1]{\left\lVert#1\right\rVert}

\title{Tweets' popularity dynamics}

\author{Ferdinand Willemin}
\email{ferdinand.willemin@ecl20.ec-lyon.fr}

\affiliation{Sahar \cite{site_sahar}, France}

\published{28th August 2022}

\keywords{IA, TIME-SERIES, TWITTER, VIRALITY, CLUSTERING, HDBSCAN, BIG DATA, TV DENOISING} 

\begin{document}

\begin{abstract}

This article charts the work of a 4 month project aimed at automatically identifying patterns of tweets' popularity evolution using Machine Learning and Deep Learning techniques. To apprehend both the data and the extent of the problem, a straightforward clustering algorithm based on a point to point distance is used. Then, in an attempt to refine the algorithm, various analyses especially using feature extraction techniques are conducted. Although the algorithm eventually fails to automate such a task, this exercise raises a complex but necessary issue touching on the impact of virality on social networks. 

\end{abstract}


\section{INTRODUCTION}
\label{sec:int}

\subsection{Context and stakes}
\label{subsec:context}

Sahar is a private company specialized in collecting, processing and visualizing massive open-access data available in the web, including social networks data. The firm provides a data analysis tool tailored to very different types of clients, requiring it to be both \textbf{exhaustive} and \textbf{flexible}. \\

This work aims to \textbf{comprehend popularity mechanisms within social networks} in an attempt to help improve some of Sahar's tool features.  \\

Nowadays, Twitter is the \textbf{ultimate medium for information broadcasting}. It encompasses more than 200 millions users\footnote{\url{https://www.blogdumoderateur.com/chiffres-twitter/}} around the world and is likely to serve both individuals as well as institutions. Moreover, its recent effects over politics and the economy have placed it at the core of global attention. This is why we picked it to study popularity mechanisms on social networks.

\subsection{Twitter}
\label{subsec:twitter}

Twitter \cite{site_twitter} is a social network, and more specifically a \textbf{micro-blogging service}, where users can post and interact with messages known as \textit{tweets}. A user can interact with a tweet in three  different ways : he can \textit{comment} it, \textit{retweet} it or \textit{like} it. To comment a tweet is to publicly answer a tweet, with a message that will appear in the tweet's \textit{comment section}. To retweet means to display a tweet on one's own profile, in order to share it with one's \textit{followers} (people that are aware of your activity on the network). To like is to show one's consensus and/or support to a message. Tweets are limited to 280 characters which makes them as much  \textbf{easily broadcastable} as \textbf{highly ephemeral}. Tweets often come with \textit{hashtags} : keywords preceded by the typographic sign \# added with the aim of specifying themes the tweet resonates to.\\ 

\begin{figure} [H]
    \centering
    \includegraphics[width=\textwidth]{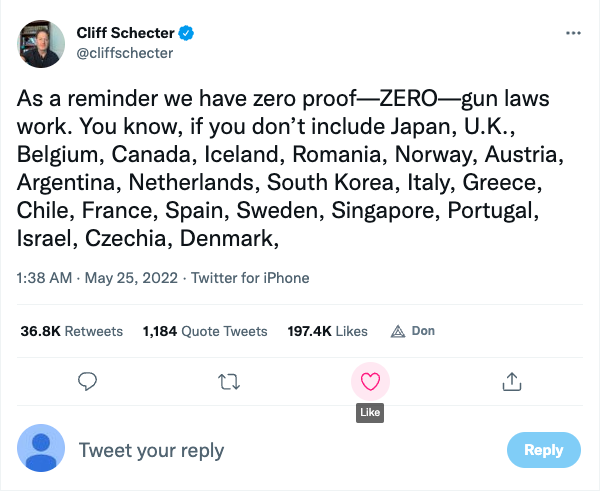}
    \caption{Screen shot of a tweet and its interaction buttons. \cite{tweet_example}}
    \label{fig:tweet_example}
\end{figure}

The manner in which  Tweets spread across the network — i.e. their way to be shared or seen by users — is atypical. That is why a study dedicated to them is conducted.

\subsection{Definition of the problem}
\label{subsec:problem}

The problem can be formulated as follows : \textit{Can a \textbf{reasonable} number of \textbf{interpretable} patterns showing the dynamics of \textbf{tweets' popoularity} through time be identified ? } \\

In order to provide a sufficient answer to the above problematic, certain subjective terms must be defined and precised. The exploratory (vs. being solely results-driven) nature of this work has opened to door to the following precisions: the \textbf{popularity} is arbitrarily designated by the \textbf{number of likes}, a \textbf{reasonable} number is likely to be \textbf{below 20} and \textbf{ interpretable} patterns must be distinct enough so that we can \textbf{name and describe them unambiguously}. \\

\noindent \textbf{Note :} In this work, we treat \textit{popularity} and \textit{virality} to be the strictly identical.

\subsection{State of the art}
\label{subsec:state_of_the_art}

Research on this topic emanates from the rise of social networks which is a relatively \textbf{recent phenomenon}. The year 2008 can be considered the representative year of this tendency, as it constitutes Facebook's passing over 100 million users \cite{facebook_history}, making it the first social network at the time. Twitter reached this count in 2011 \cite{twitter_history}.  \\

The first noticeable study dealing with popularity dynamics in user-generated content \cite{crane_sornette_2008} is applied to the video host YouTube, with the aim of finding \textit{locally relevant} content differing from the well-known \textit{most popular} content that only considers  mass appeal and an instantaneous vision. The analysis leads to the identification of three categories : the \textbf{junk}, the \textbf{quality} and the \textbf{viral} video dynamics. Junk videos undergo a burst of popularity which drops quickly afterwards because they do not spread through the social network. Quality videos meet a very sudden peak in popularity, certainly caused by an \textit{exogenous} effect (such as being featured on the first page of YouTube), and a subsequently passive decline. Viral videos face a slowly increase up to a peak followed by a slow decrease which reflects a \textit{word-of-mouth} process. Concurrently, most of the videos do not experience any peak in popularity, embodying a fourth category : the \textbf{silent} videos.

YouTube is actually not a social network but some parallels can be drawn between its features and Twitter's. For instance, the social network displays, alike Youtube, the trendiest topics on its first page which can trigger exogenous effects as well.

However, apart from the difference of contents' nature between the two websites, the article \cite{crane_sornette_2008} is based on applying a mathematical model (the \textit{self-excited Hawkes conditional Poisson process}) to the data. The assumptions made to use it are not detailed enough and may not fit with our case. As a result, the origins of the four categories are not fully defined. \\

Another interesting work \cite{yang_leskovec_2011} implements a \textbf{clustering algorithm} to gather hashtags' popularity dynamics with similar \textit{shapes}. The popularity is measured by the number of appearances of a hashtag over time. In summary, it uses a K-Means algorithm provided by an "improved" euclidean distance that ignores both the overall hashtags' appearances and the temporal gap between the popularity major peaks. 

\begin{figure} [H]
    \centering
    \includegraphics[width=\textwidth]{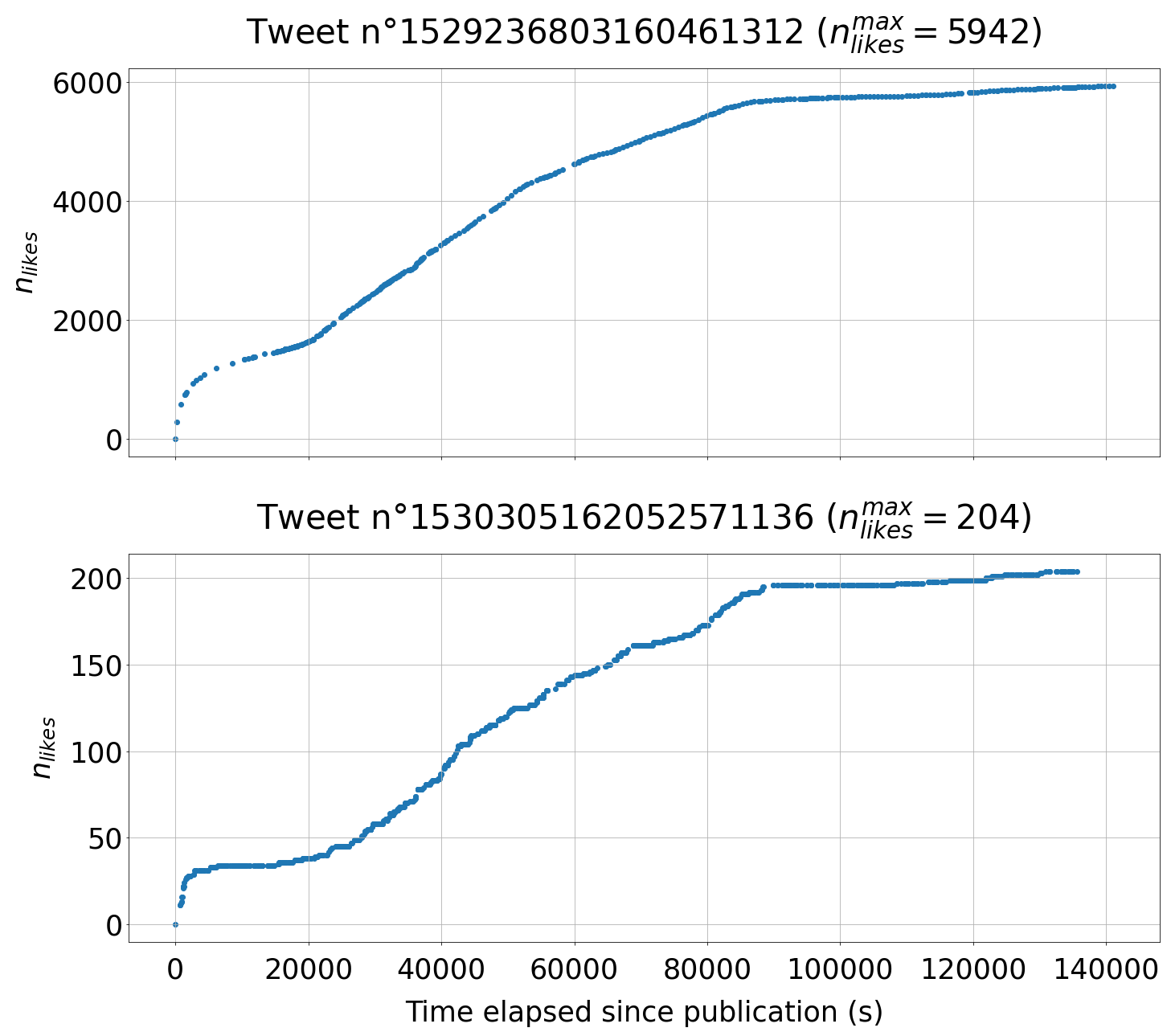}
    \caption{Two tweet's history of likes designated by their Twitter id.}
    \label{fig:same_shapes_different_dynamics}
\end{figure}

Although the idea of using a clustering algorithm may seem appropriate — since such algorithms are used to classify objects according to specified features — some of the choices made by the author hide some conjectures that are arguable in the context of our work. First, given two tweets and the history of their number of likes, does having the same \textit{shape} but at different scales with a time lag can be deemed to be similar \textit{dynamics} ? Figure \ref{fig:same_shapes_different_dynamics} illustrates this issue by showing two evolutions whose \textbf{shapes} could be considered as \textit{similar} as they both include successively a quick growth, an effective stage, a slower but longer rise and eventually a cap. However, their \textbf{maximum number of likes} ($n^{max}_{likes}$) vary with a factor of almost 30 ! The processes that generated them do not engage the same range thus their \textbf{dynamics} are different. The use of a K-Means algorithm raises other questions, especially around the choice of the K hyperparameter which sets the number of clusters. Moreover, the preprocessing adopted —implying to manipulate the 1000 most frequently mentioned hashtags — does not guarantee to lead to a large enough dataset where all patterns possible are represented.

Ahmed et al. \cite{ahmed_2013} also makes use of a clustering strategy. They build a two-dimensional feature space and a correlation-based similarity metric combined with the affinity propagation algorithm \cite{affinity_propagation} to extract the main evolution patterns from their datasets. Their approach is worth exploring but its complexity — especially the simultaneous analysis of two distant measures carrying information of different nature — makes it a method for further investigations that should be implemented at a later stage. Unfortunately, time did not allow our study to experiment with this model. \\

A different way of seeing things, presented by C. Li, J. Liu, and S. Ouyang \cite{li_2016}, represents the evolution of videos' popularity from the chinese service provider Youku with a succession of two states : 1 if the video is experiencing a \textit{burst} of popularity and 0 otherwise, the \textit{popularity} corresponding to the number of views in a day. A \textit{burst} stands for a striking increase in popularity. Quantitatively, it is represented by an \textbf{exceeding of a certain threshold by the derivative}, whose value is actually established at \textbf{three times the average derivative}. The idea of extracting key information from curves to discriminate them is encouraging but requires additional work to be carried out in order to gain in rigor. 

Among others, this representation may \textbf{only embrace sketchy tendencies} and by doing so neglect some nuances. For instance, integrating the size or duration of the different states could allow us to distinguish subgroups between them or might reveal some typical behaviors related to them. More specifically, the threshold's value must come with a \textbf{reliable justification} that exhibit a true social phenomenon and the \textbf{burst's mathematical definition} itself may need to be reviewed to improve its robustness.  \\

All in all, many lines of attack have been exposed over the previous years. They should now be used to meticulously define a tailor-made strategy which will attempt to answer the initial problem.

\subsection{Organization of the article}
\label{subsec:organization}

Once \textbf{the data presented} (\ref{sec:data_presentation}), we will implement a \textbf{HDBSCAN clustering algorithm} \cite{hdbscan_library} specifically designed for our study (\ref{sec:clustering}). Then, in an attempt to produce customizable and better results, another version of the HDBSCAN algorithm will be applied on a transformed dataset (\ref{sec:tweet_vector}). By doing so, an original method created to \textbf{reduce and store information} of the popularity's history into what is called a \textit{tweet vector} will be introduced. Ultimately, an \textbf{overall assessment} of the techniques developed followed by a \textbf{list of promising avenues} for further exploration of the topic will be proposed (\ref{sec:retrospection}).

\section{PRESENTATION OF THE DATA}
\label{sec:data_presentation}

\subsection{The data fetching system}
\label{subsec:scrapping_feature}

Part of Sahar's solution is an intelligent \textit{web scraping}\footnote{Act of collecting publicly available data from the web, often on a large scale.} feature of Twitter. It notably enables users to analyze all messages containing one or several keywords — commonly a hashtag — and all the ones published by one or several given users. Thus, collecting an important chunk of messages around a particular subject is made extremely efficient and smooth. 

There are \textbf{two main} advantages to using a topic-oriented scraping. On the one hand, tracking all tweets posted in a given period of time would be \textbf{too costly both in time and energy} : around 500 millions tweets are produced each day \cite{internet_live_stats-2011} and assuming that their average \textit{lifetime}\footnote{Duration starting when the message is published and ending when it no longer generates interactions.} is a few days, each dataset would hold billions of tweets. On the other hand, it is conceivable that \textbf{evolutionary patterns will vary depending on the topics addressed}. Hence, Sahar's tool's filtering capability prevent us from mixing the topics — and therefore the patterns — too much. 

Furthermore, let us recall that the project's conditions (\ref{subsec:context}) demand the virality algorithm to be \textbf{compatible} with the other features of Sahar's tool, as they are to work in synergy. From all the above, it only seems natural that we use the company's own scrapping method. \\

To ensure a significant \textbf{diversity} regarding the amplitude of the data monitored, the keywords or accounts followed were selected for their \textbf{capacity to generate many likes}. Without this precision, we would automatically collect all tweets, especially noisy ones that do not generate any interaction. In effect, these \textit{inert tweets} constitute the large majority of the publications \cite{huffpost_2010}. Besides, by removing those inert tweets not only do we save time, but we also prevent our datasets from being excessively large.

\subsection{The resulting datasets}
\label{subsec:datasets}

\textbf{Two datasets} are used in this study. The first one, named \textit{dataset 1}, is composed of \textbf{2785 tweets} observed from 23/05/2022 to 27/05/2022 included. The points of each time series are recorded every 10 minutes.
The second one, \textit{dataset 2}, is composed of \textbf{3000 tweets} monitored from 17/06/2022 to 26/06/2022 included with a record every 5 minutes. The keywords and users associated to each one of them are detailed in \ref{appendix:datasets}. \\

\noindent N.B: It must be kept in mind that the frequency of data acquisition is not always constant in practice, due to the web scrapping feature performances.

\subsection{Rudimentary preprocessing}
\label{subsec:rudimentary_preprocessing}

\begin{figure} 
    \centering
    \includegraphics[width=\textwidth]{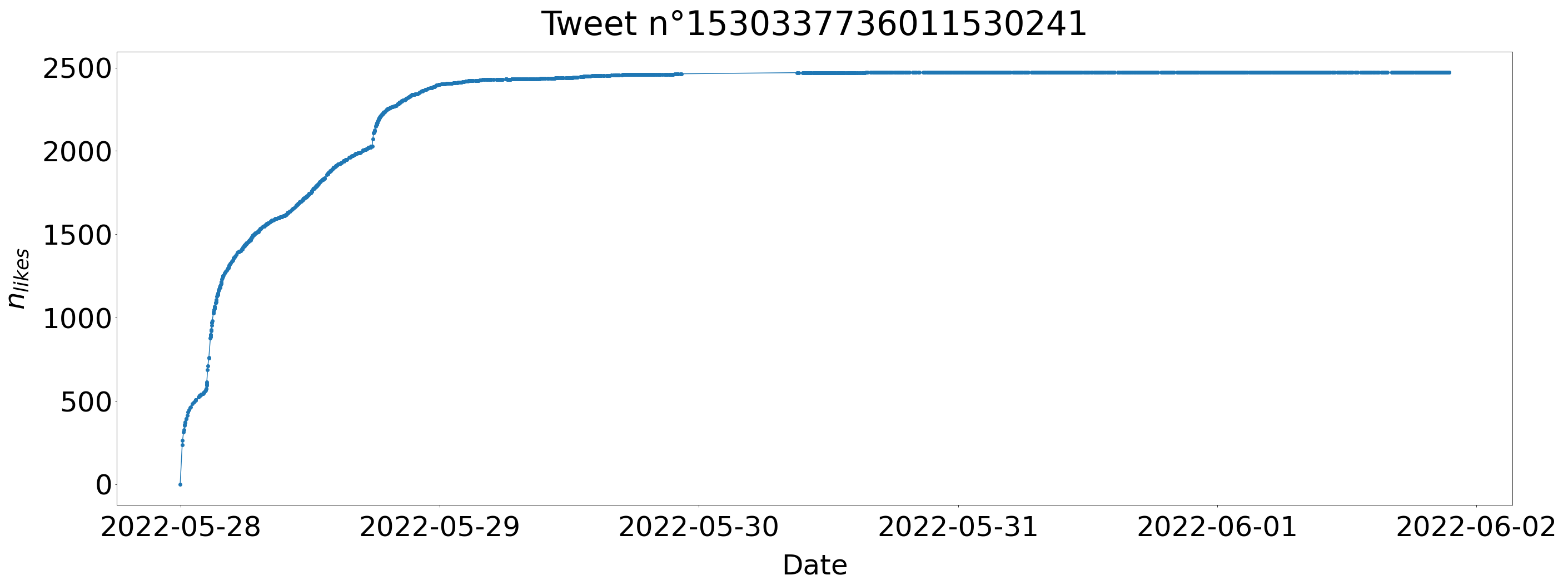}
    \caption{Typical raw data from the $1^{st}$ dataset.}
    \label{fig:raw_data}
\end{figure}

Figure \ref{fig:raw_data} shows the evolution of popularity right after it had been collected. The exact time is not specified to not overload the plot although it has a 1-second accuracy.

\begin{figure}
    \centering
    \includegraphics[width=\textwidth]{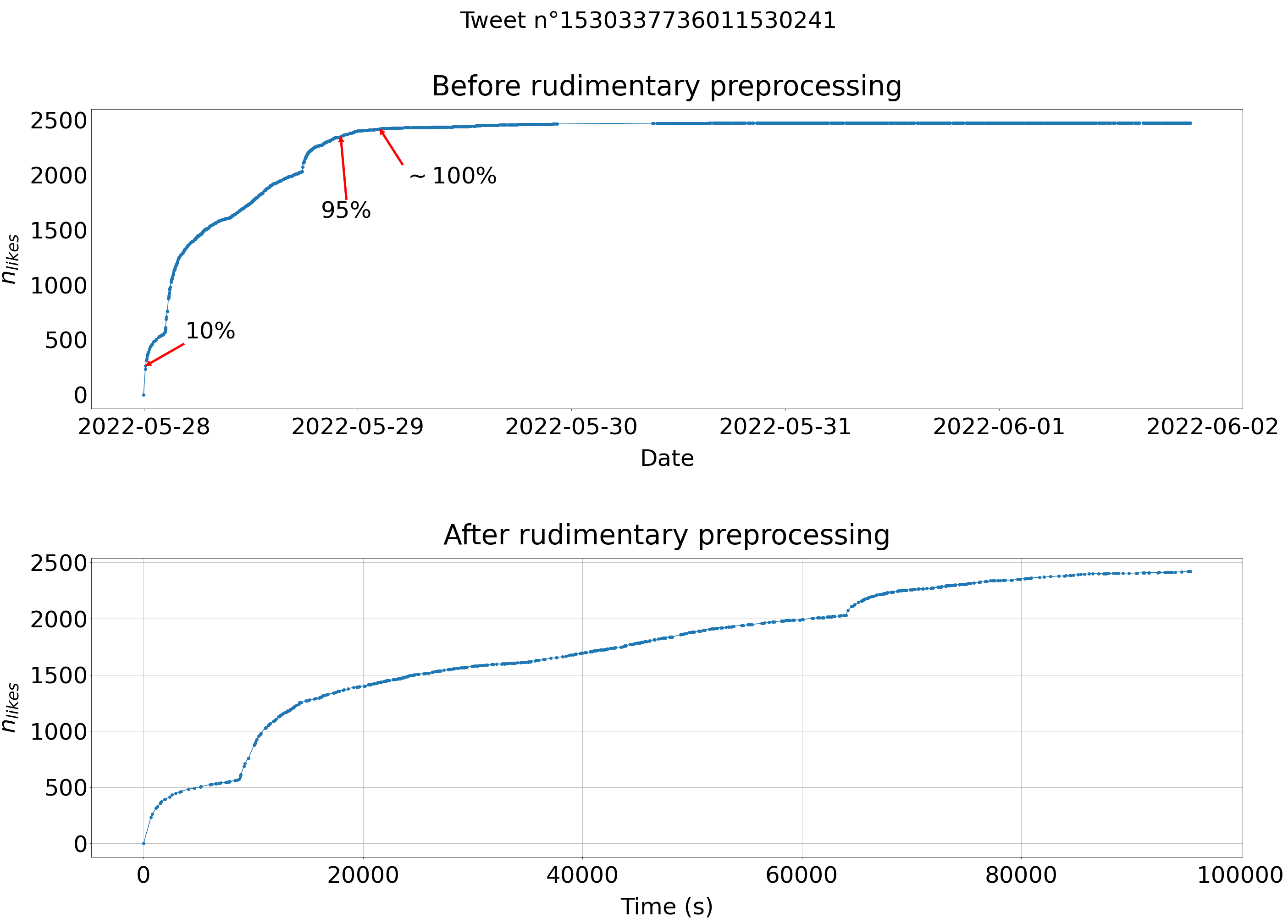}
    \caption{Example of the raw data and its elementary values (on top) transformed into a more suitable format (on the bottom).}
    \label{fig:rudimentary_preprocessing}
\end{figure}

This raw data format must undergo a first and obvious \textbf{preprocessing} (Figure \ref{fig:rudimentary_preprocessing}). To enable \textbf{comparison} between the different evolutions, the x axis is expressed in terms of \textit{duration} in SI units. In the meantime, a large part of the asymptotic behavior is removed since it doesn't carry any relevant information. To do so, we store the values $t_{10}$ and $t_{95}$ (respectively the times at which the curve reach 10\% and 95\% of its maximum value). This draws an interval $\Delta t=t_{95}-t_{10}$. Then, we "extend" the interval by 20\% by defining $t_{max}=t_{10}+1,2.\Delta t$. Finally, the data from 0 to $t_{max}$ provides a \textit{window} \textbf{focused on the evolution dynamics}.

\section{CLUSTERING}
\label{sec:clustering}

Clustering belongs to the \textit{unsupervised} class of machine learning algorithm because it \textbf{doesn't involve two steps} including a \textbf{training stage} where inputs are entered simultaneously as their corresponding outputs. On the contrary, it engages a \textbf{single step} where inputs are agglomerated to form clusters according to their \textit{distance} to each other. Therefore, building a metric which represents \textbf{how close the popularity evolutions are} is fundamental. Because it truely influences the quality of the results, it is the touchy phase of the process.

To design such a model, it is essential to answer the following questions beforehand : what does \textit{having similar popularity dynamics} means ? and what is a \textit{good} cluster in our case ? Through a naive approach, we answer the first problem by looking for a \textbf{point to point proximity}. The second issue is more tricky since it implies \textbf{subjective criteria}. For instance, one may prefer to get homogeneous clusters regarding the number of tweets it contains without being demanding on the groups' inner proximity, while another may want to obtain very accurate clusters, whatever the quantity of curves they hold. To get around such complications, \textbf{quality is arbitrarily favored}, taking into account that our notion of proximity is by nature very demanding.

In addition, the algorithm chosen in this study comes with a useful feature : the existence of a \textit{noise cluster}, where all the unique dynamics — i.e. those that are "far" to every other — are placed. The \textbf{size} of this extra lot is also a criteria that can lead to new compromises.
 
\subsection{Choice of the metric}
\label{subsec:metric}

A prior search of the existing work conducted to the testing of \textbf{two} different distances : the \textit{Dynamic Time Warping distance} and what we call the \textit{L1 distance}. The first one naturally allows to avoid the main problem caused by the inaccurate recording of the data (cf \ref{subsec:datasets}) : the \textbf{disparity of the values along the time axis}. The L1 distance doesn't possess the same ability so a \textbf{more elaborated preprocessing} is needed to tackle this issue. However, the latter is \textbf{faster to compute} and \textbf{more respectful} of the temporal distribution.

\subsubsection{Dynamic Time Warping distance}
\label{subsubsec:dtw}

\begin{figure}
    \centering
    \includegraphics[width=\textwidth]{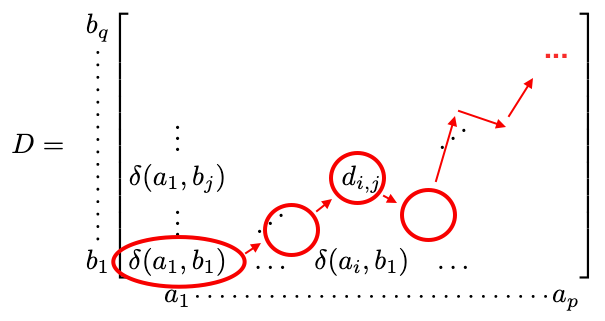}
    \caption{Visualization of a fictional distance matrix and its warping path (in red).}
    \label{fig:warping_path}
\end{figure}

The Dynamic Time Warping distance (or DTW distance) between two time series A and B is introduced in \cite{berndt_1994}. It is determined as follow :

\noindent Let $\{a_1, \ldots, a_p\}$ and $\{b_1, \ldots, b_q\}$ respectively be the values of A and B \textbf{along the Y axis} (note that p and q don't have to be equal). Then, let W designate the set of all the \textit{paths} between A and B, i.e. the selections of $(i, j)$ where every points from A and B are involved at least once. Let also consider that a simple distance $\delta$, typically the euclidean distance, has been settled. For all $i \in \{1, \ldots, p\}$ and $j \in \{1, \ldots, q\}$, $\delta(a_i, b_j)$ is calculated. The results can be visualized in a \textit{distance matrix} 
$D=\left(d_{i, j}\right)_{1 \leq i \leq p \atop 1 \leq j \leq q}$ 
where $d_{i, j} = \delta(a_i, b_j)$ (figure \ref{fig:warping_path}). The \textit{DTW distance} is then defined as 

\begin{equation}
    DTW(A,B) = \displaystyle\min_{w \in W}\left(\sum_{(i, j) \in w} \delta(a_i, b_j) \right)
\end{equation}

\begin{figure}
    \centering
    \includegraphics[width=\textwidth]{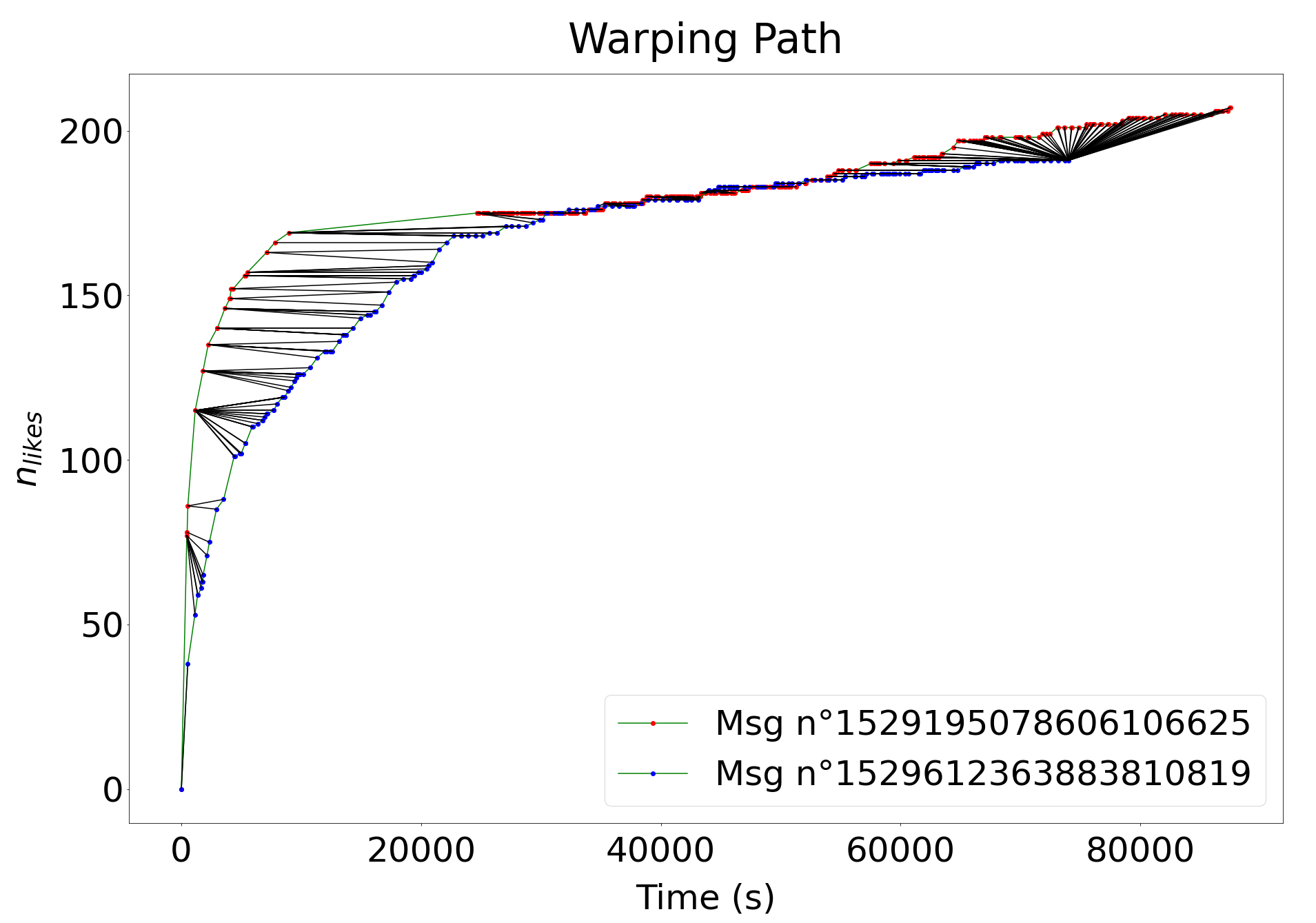}
    \caption{Illustration of the warping path (black lines) between two tweets.}
    \label{fig:warping_path_viz}
\end{figure}

\noindent The path $w$ that lead to the DTW distance is called the \textit{warping path} (see figures \ref{fig:warping_path} and \ref{fig:warping_path_viz}). \\

This is equivalent to applying a \textit{non-linear temporal distortion to the data} which aligns on the time axis points that are the closest towards the Y axis and then return the sum of their distances.

 Many aspects of this definition are \textbf{physically incoherent} : for a given i, calculating $\delta(a_i, b_j) \: \forall j$ means that the temporal dimension is neglected. To address these obstacles, some \textbf{correcting constraints} are added :
\begin{itemize}
	\item the \textit{boundary constraint} stipulates that a path must include (1, 1) and (p, q) which guarantees to consider a \textbf{beginning} and \textbf{end} notion ;
	\item the \textit{monotonicity constraint} states that $i_k \leq i_{k+1} \: \forall k$ and $j_k \leq j_{k+1} \: \forall k$ which compels the points inspection to always \textbf{advance in time} ;
	\item the \textit{continuity constraint} imposes that $|i_{k+1}-i_k| \leq 1 \: \forall k$ and $|j_{k+1}-j_k| \leq 1 \: \forall k$ which forces comparing distances between \textbf{neighbor points}, and therefore to progress in both time series \textbf{"at the same speed"} ;
	\item the \textit{warping window} specifies a \textbf{maximum range (R)} of points to visit to prevent the points scan to be \textbf{"stuck"} in one time series while it keep going in the other : it can be expressed as $\forall k, |i_k-j_k| \leq R$.
\end{itemize}

\begin{figure}
    \centering
    \includegraphics[width=\textwidth]{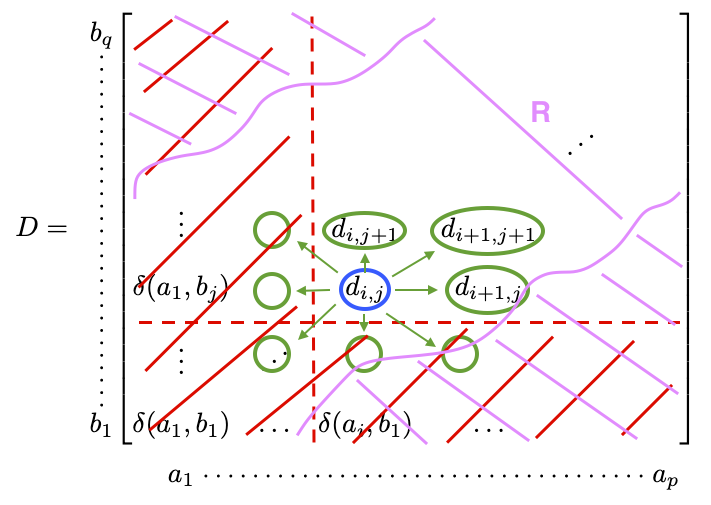}
    \caption{Illustration of the DTW correcting constraints and their effect on the algorithm : the \textit{monotonicity constraint} in red, the \textit{continuity constraint} in green and the \textit{warping window} in purple. Here, the path can only continue to the green locations that are not in hatched zones.}
    \label{fig:dtw_constraints}
\end{figure}

Once these restrictions are employed, a recursive formula can be revealed to find a suitable warping path :

\begin{equation}
    \gamma(i,j)=\delta(i,j)+min[\gamma(i-1,j),\gamma(i-1,j-1),\gamma(i,j-1)]
\end{equation}
\noindent where $\gamma(i, j)$ stands for the cumulative distance until point (i, j). \\

\cite{berndt_1994} also proposes an algorithm to determine this distance — the \textit{DTW algorithm} — which runs in \textbf{quadratic time} ($O(pq)$). However, an optimization of the correcting constraints combined with an appropriate reduction of the data (called \textit{Abstraction approach}) can result in an upgrade that works in \textbf{linear time} : the \textit{FastDTW} \cite{fast_DTW}. The latter is used for our calculations.

\subsubsection{"L1" Distance}
\label{subsubsec:L1_distance}

A more straightforward method consists in computing a point to point distance between the curves. However, comparing them at similar instants requires to have their points \textbf{aligned along the time axis} and to carry the \textbf{same amount of points}, otherwise the distance cannot be determined. As mentioned before (section \ref{subsec:metric}), our data doesn't satisfy these conditions. \\

To overpass this issue, the curves must be \textbf{interpolated linearly with a fixed step} \cite{linear_interpolation_wiki}. To limit the loss of information caused by this process the step is set at 5min, i.e. \textbf{twice as small as our recording frequency}. To ensure that they all possess the same number of points, they are \textbf{extended by considering they have reached their asymptote} : points equal to the last existing one are added until they get to the longest evolution.\footnote{\textbf{Note :} With the benefit of hindsight, it would have been easier and more legitimate to modify the rudimentary preprocessing (section \ref{subsec:rudimentary_preprocessing}) by truncating the asymptotes at the \textbf{maximum $t_{max}$ of the dataset}.} \\

Conclusively, given $n \in \mathbb{N}^*$ and two interpolated time series A and B with their respective points $\{a_1, \ldots, a_n\}$ and $\{b_1, \ldots, b_n\}$, the distance is expressed by :

\begin{equation}
    d_{L1} = \sum_{i=1}^n |a_i - b_i|
    \label{eq:d_L1}
\end{equation}

\begin{figure}
    \centering
    \includegraphics[width=\textwidth]{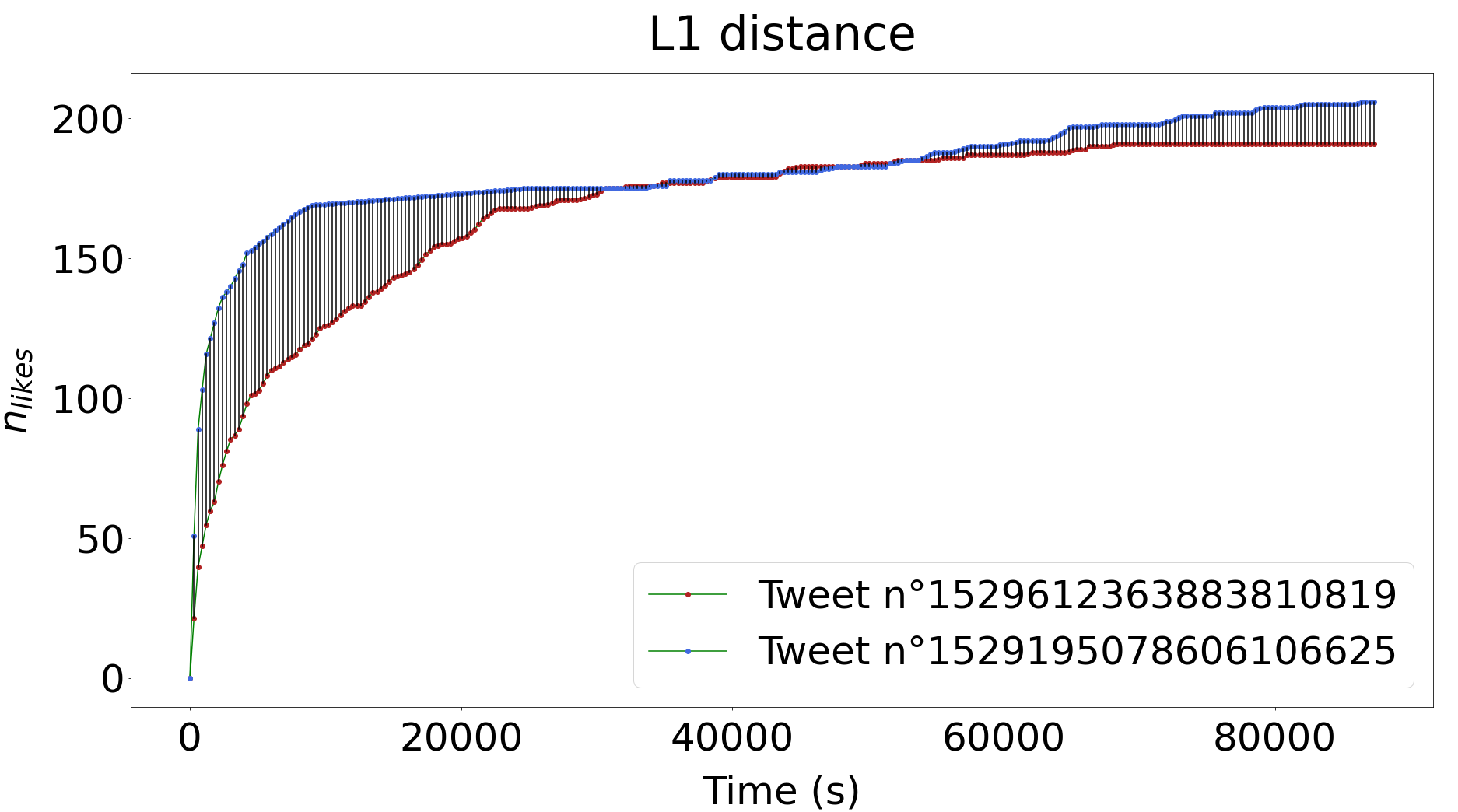}
    \caption{Illustration of the L1 distance between two tweets.}
    \label{fig:L1_distance_viz}
\end{figure}

\noindent In its continuous form (i.e. when step $\to 0$), (\ref{eq:d_L1}) is written $d_{L1}=\int\limits_0^{t_{max}}|f-g|dx$ where $f$ and $g$ are the functions describing the two evolutions, which inspired the name \textit{L1 distance}. It coincides with the surface visible between the two curves (figure \ref{fig:L1_distance_viz}). The lower this area is, the closer the dynamics are. 

\subsubsection{Analysis of the distances}
\label{subsubsec:distance_analysis}

Discriminating the metrics isn't trivial. Indeed, their main goal is to bring the equivalent curves closer together and to move away from each other those that are different, so they must be judged on this task. However, they were built for this very purpose. The fact that \textbf{they serve as both a test object and an evaluation instrument} forces us to imagine alternative criteria of performance. \\

Knowing that the human eyes are the best tools to execute this task for a low number of tweets, a \textbf{qualitative survey} is first conducted. For each distance the $i^{th}$ closest pair of curves is examined and compared ($i$ taking different values). Besides, some \textbf{test samples} are selected. Each one contains two pairs of tweets : one considered as \textit{close} and the other as a \textit{distant} pair according to human eyes. The goal is to see whether the pairs are \textbf{labeled identically} with our metrics. To make it happen, a \textbf{random sub-sampling} of the $1^{st}$ dataset is implemented which selects \textbf{200 tweets} out of 3278. This is necessary because \textbf{the DTW algorithm is quite time consuming}. As a reference, it takes about \textbf{5 seconds} to calculate the pairwise L1 distance between the sub-sampled data (including the interpolation step) while it takes about \textbf{30 minutes} with the DTW distance. Incidentally, this facet counts as a \textbf{great prejudice} against the latter.

\begin{figure*} [hbt!]
    \centering
    \includegraphics[width=0.45\textwidth]{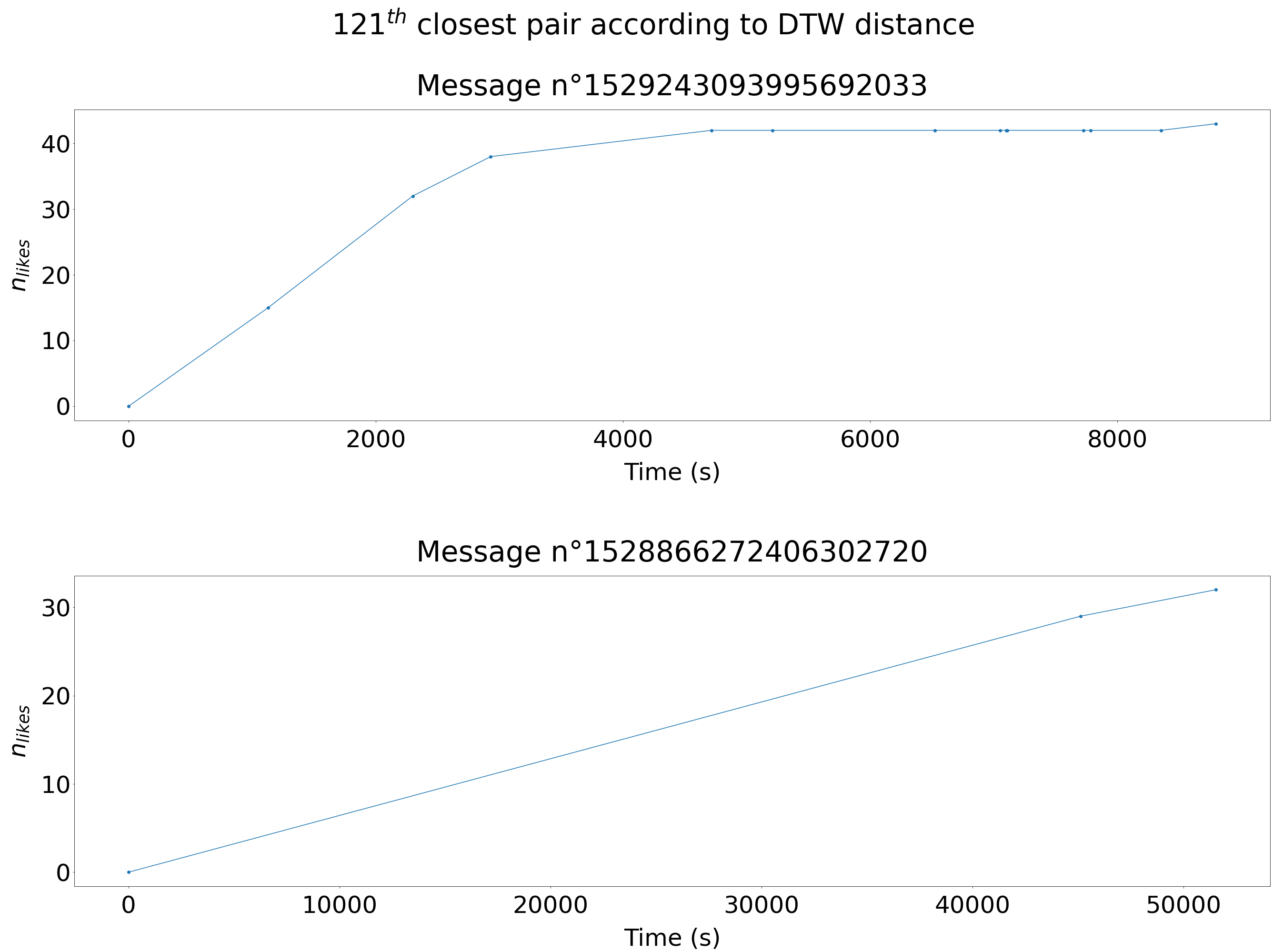}
    \includegraphics[width=0.45\textwidth]{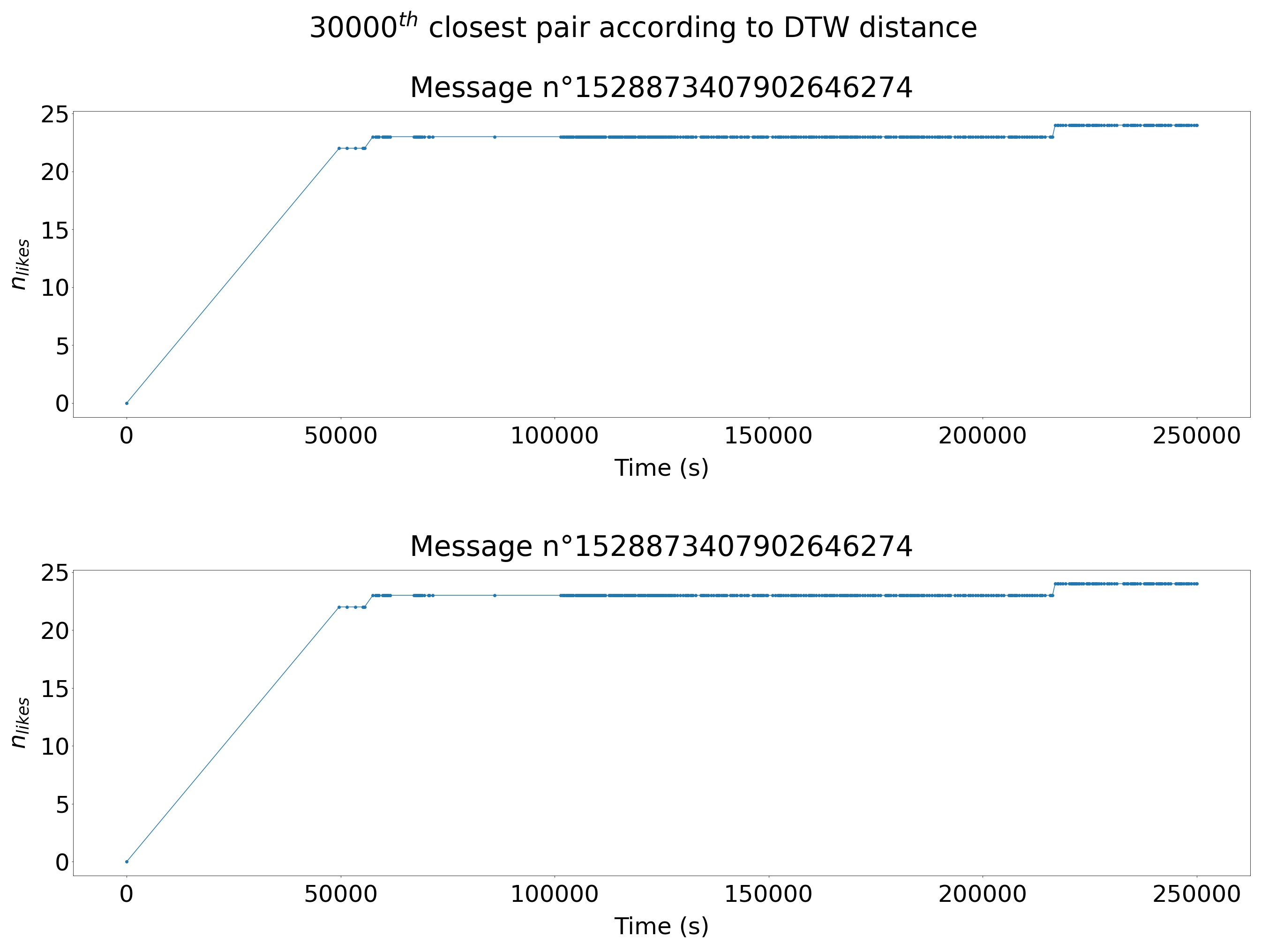}
    \caption{Two pairs of tweets extracted and their "proximity rank". According to the DTW distance, the left pair contains \textbf{closer} tweets than in the right one. Qualitatively, one can judge the opposite.}
    \label{fig:dtw_inaccuracy}
\end{figure*}

This inquiry reveals some weaknesses of DTW, as it sometimes leads to doubtful results (figure \ref{fig:dtw_inaccuracy}). On the contrary, the L1 distance does not experience such outcomes. \\

\begin{figure*} [hbt!]
    \centering
    \includegraphics[width=0.45\textwidth]{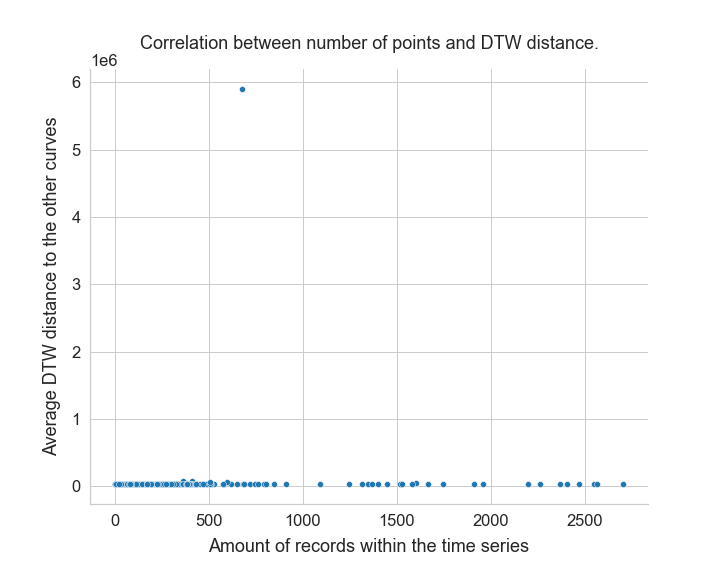}
    \includegraphics[width=0.45\textwidth]{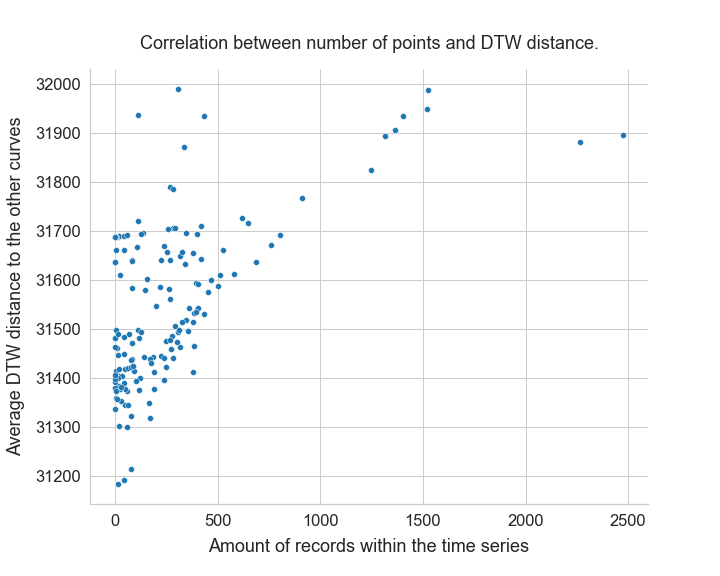}
    \caption{Variation of the average DTW distance according to the number of points (raw plot on the left, without extreme points on the right). It seems that time series carrying a lot of information (i.e. many records) have a tendency to move away from the others relatively to the DTW distance.}
    \label{fig:corr_nb-points_dtw}
\end{figure*}

Even if a qualitative examination is not sufficient to conclude, it exhibits another proof of DTW's inferiority. As shown in figure \ref{fig:corr_nb-points_dtw}, this distance is slightly influenced by the number of records present in the time series. Despite some attempts to get rid of this drawback (mainly through a penalization strategy), it remains disturbing. Bypassing the problem with preprocessing would be absurd since it constitutes the main strength of the algorithm. \\

All in all, due to its better \textbf{robustness}, \textbf{simplicity} and \textbf{speed}, the L1 distance is chosen as the reference distance for the rest of the project. However, DTW is worth exploring. Its inherent upsides may be useful for other applications and can be greatly improved in other contexts. As for the computational time, one must notice that the algorithms used may not be fully optimized.

\subsubsection{Adding a weighting}
\label{subsubsec:weighting_distance}

Afterwards, it has been assumed that the \textbf{beginning} of the popularity evolution was carrying \textbf{more information about the underlying social dynamics} than the end. To introduce this feature into our metric, a weighting is added, \textbf{favoring the first instants} and \textbf{handicapping the last ones}. \\

Concretely, it implies multiplying the L1 distance with a \textbf{penalty function}. Several functions were studied to eventually select the one that offered the \textbf{most control}. It is defined as :

\begin{equation}
    f(t)=(1 - \epsilon) * \left(\frac{th(\beta-\alpha.t)+1}{2}\right) + \epsilon
\end{equation}

\noindent where :
\begin{itemize}
    \item $\beta$ is such that $f(0)=0,99$ ;
    \item $\alpha$ is such as $f(\overline{t_{max}}) = 0.7$ with $\overline{t_{max}}$ being the median of all the $t_{max}$ of the dataset ;
    \item $\epsilon=0.05$ stipulating that the minimum weight is 5\%
\end{itemize}

\begin{figure} [H]
    \centering
    \includegraphics[width=\textwidth]{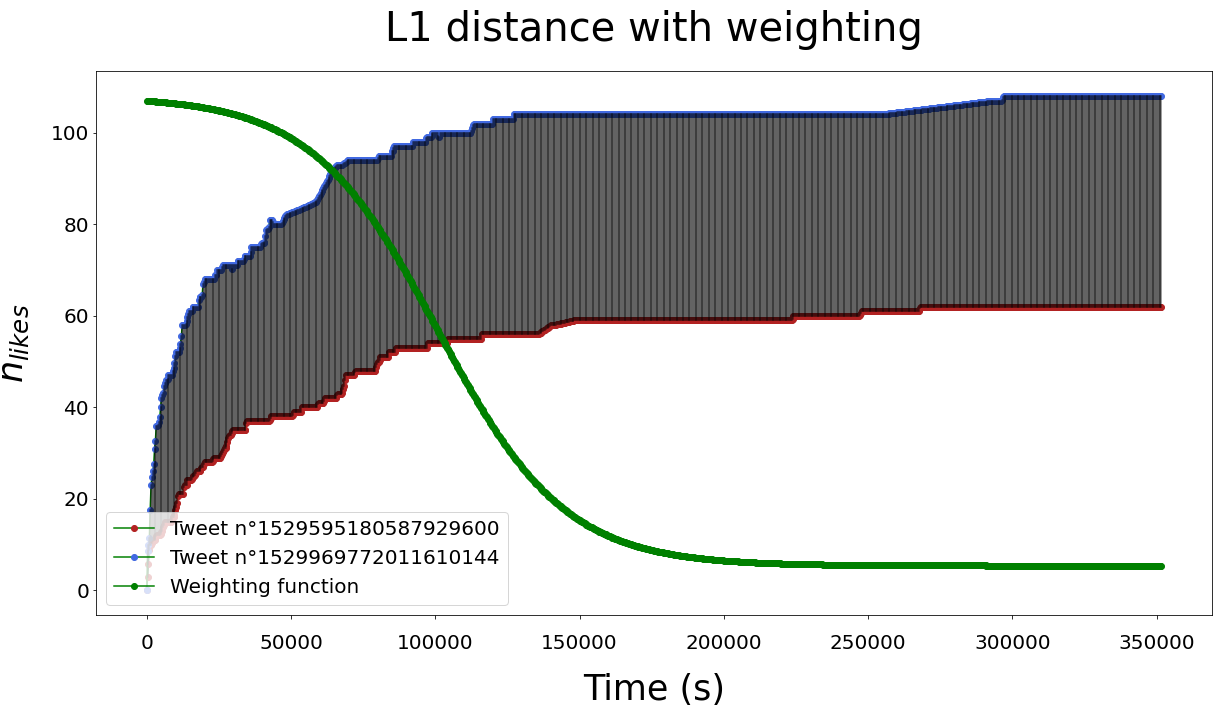}
    \caption{Graph showing the impact of the weighting on the L1 distance. \textbf{Note} : the amplitude of the weighting has been increased for the figure.}
    \label{fig:L1_distance_viz_with_weighting}
\end{figure}

\noindent The weighting's influence can be seized in figure \ref{fig:L1_distance_viz_with_weighting}.

\subsection{The clustering algorithm}
\label{subsec:clustering_algorithm}

\subsubsection{HDBSCAN}
\label{subsubsec:hdbscan}

HDBSCAN \cite{campello_density-based_2013} is originally a DBSCAN \cite{dbscan} upgrade. Both are \textbf{density-based} algorithm, which means they have been conceived to identify dense areas within groups of points. As mentioned before, one of their considerable advantages is their capacity to generate a \textit{noise cluster} where all the unique dynamics are stored. Actually, all time series that \textbf{are not in a dense zone} are considered as noise. \\

To execute such a task, it needs to be provided with two main parameters : \mintinline{python}{min_samples} and \mintinline{python}{min_cluster_size}. \\

\mintinline{python}{min_samples} is used to \textbf{define a disk surrounding each data point} whose ray corresponds to the \textbf{distance between a point and its} \mintinline{python}{min_samples}\textbf{'s closest neighbor}. Two points whose L2 distance is inferior to both point's rays will be considered to be in the \textbf{same dense area}. If only \textbf{one of the two} belongs to the disk of the other, their distance will be increased to match with the bigger ray. Therefore, \mintinline{python}{min_samples} is used to \textbf{redefine the space topology by accentuating density phenomena}: the points which are not in dense areas are even more isolated. The distance created by this process is called the \textit{mutual reachability distance}. \\

\mintinline{python}{min_cluster_size} corresponds to the \textbf{minimum size for a dense area to be considered as a cluster}. Bellow this value the time series inside the zone are viewed as \textbf{noise}. To better seize its impact, a deeper comprehension of the algorithm is needed. 

\begin{figure} [H]
    \centering
    \includegraphics[width=0.95\textwidth]{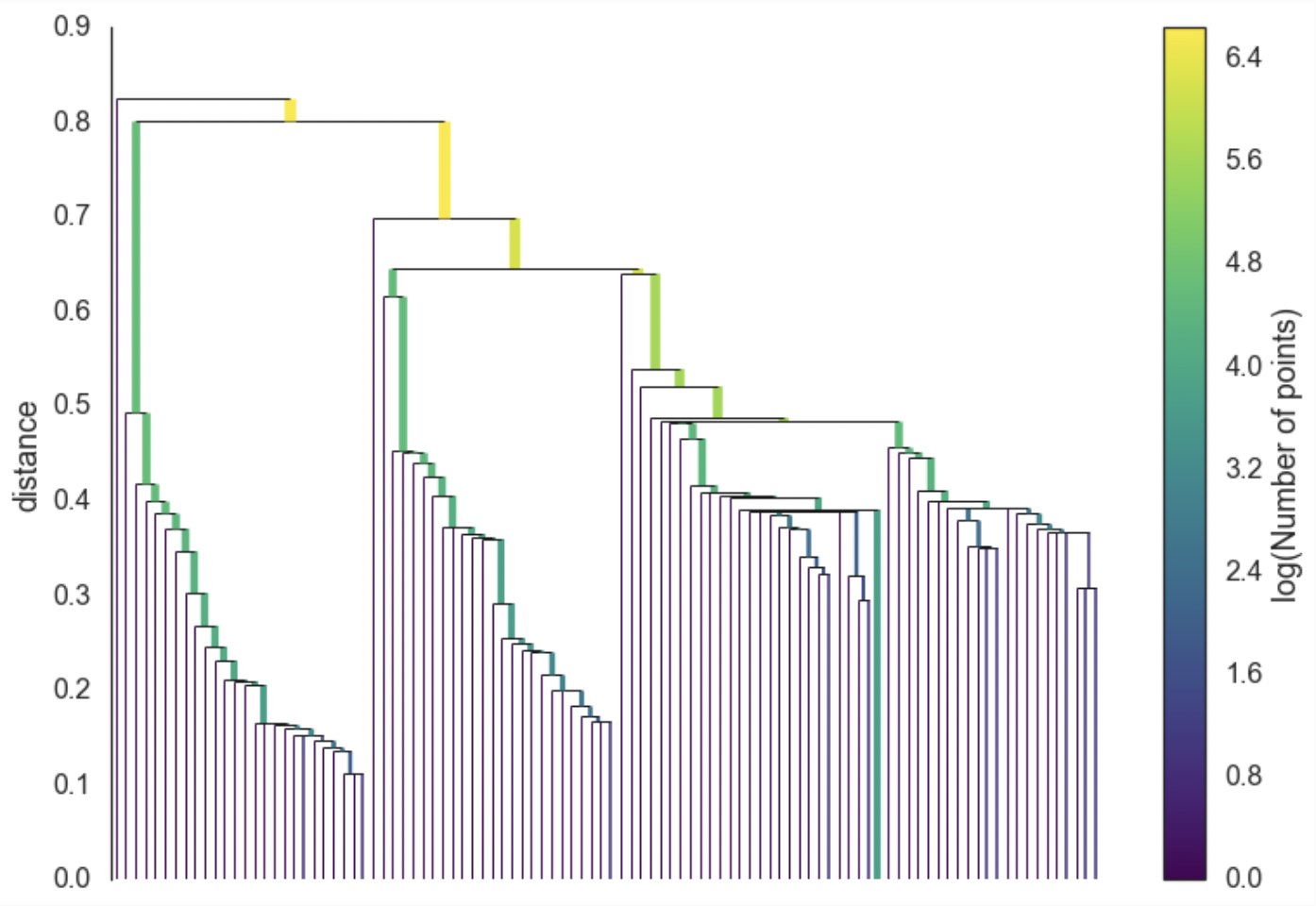}
    \caption{Dendogram depicting clusters' splits as $\epsilon$ (here being normalized and called \textit{distance}) decreases. \cite{hdbscan_library}}
    \label{fig:hdbscan_dendogram}
\end{figure}

Let's consider the minimum spanning tree of the dataset \cite{minimum_spanning_tree} built from the mutual reachability distance. By definition, this tree links all the points together each time with a single bound. Let's consider that when a point is linked with another, they both form a cluster. Let's delete the bounds one by one from the biggest to the smallest. Removing a bound is equivalent to divide a parent cluster into two children clusters. Therefore at the end, all the points are isolated from the others. All the divisions triggered during the process can be visualized in a dendogram as in figure \ref{fig:hdbscan_dendogram}.

\begin{figure} [H]
    \centering
    \includegraphics[width=\textwidth]{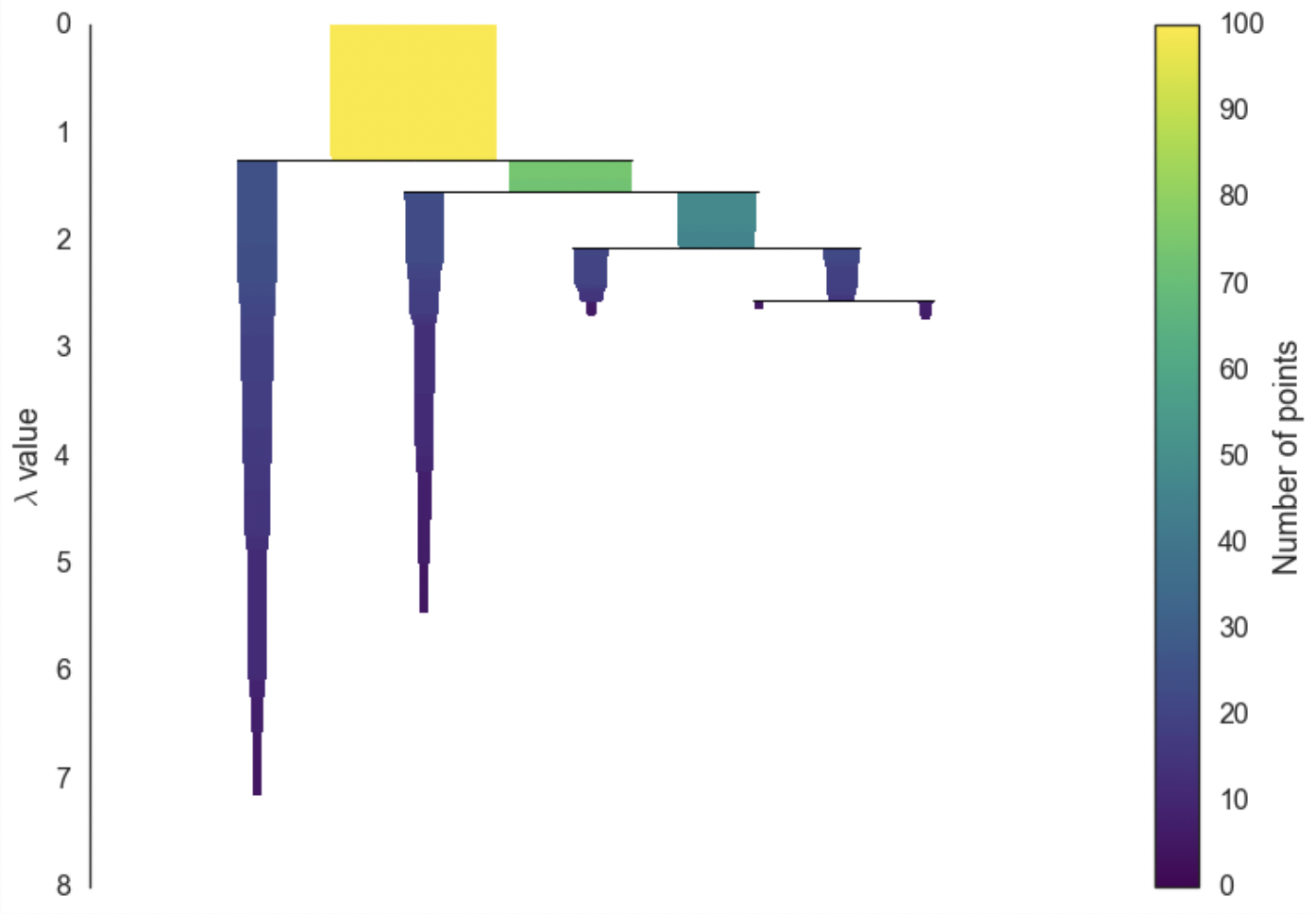}
    \caption{The same dendogram after applying \mintinline{python}{min_cluster_size} $\left(\lambda=\frac{1}{\epsilon}\right)$ \cite{hdbscan_library}}
    \label{fig:hdbscan_dendogram_erosion}
\end{figure}

Since this method leads to an enormous amount of clusters, a refinement is added : rather than seeing each split as a parent cluster giving birth to two children clusters, it could be interpreted as a \textbf{"cluster erosion"} : if a children is \textit{too small} to be a cluster itself, it means that the parent cluster didn't actually split, but instead lost some points that became noise. Hence, \mintinline{python}{min_cluster_size} is there to \textbf{define} what \textit{too small} means. Figure \ref{fig:hdbscan_dendogram_erosion} shows the kind of change provoked on the dendogram.

Eventually, a \textit{flat clustering} is extracted by keeping the most \textit{stable} groups that don't overlap with each others. Crudely, they relate to the longest ones in the dendogram. \\

This stands for the basic knowledge necessary to understand our way of using HDBSCAN. A more exhaustive and detailed explanation can be found in the library documentation \cite{hdbscan_library}.

\subsubsection{Parameters selection}
\label{subsubsec:parameters_selection}

Building a \textbf{score} to choose the parameters in a way that guarantees both a \textbf{reasonable number} of clusters and a \textbf{sufficient proximity} within them was first intended. Unfortunately, it didn't succeed mainly because it was \textbf{too sensitive} : the score obtained after an important variation of the parameters — around 10 units — was so close that the "optimal" parameters could dramatically change between the datasets. These parameters are so affecting that such a behavior cannot be allowed. \\

As a result, another indicator is adopted : the \textbf{size of the noise cluster}. Usually, this quantity is relatively large — from 40 to 50\% of the dataset — which is quite annoying. Hence, it is desirable to \textbf{reduce it as much as possible}. The best lowering is achieved with the lowest values of both \mintinline{python}{min_cluster_size} and \mintinline{python}{min_samples}, i.e. \textbf{2}. It is not surprising at all knowing their influence on the clustering (see \ref{subsubsec:hdbscan}). Considering that the \textbf{accuracy of the clusters is favored} over a reasonable amount of groups (cf \ref{sec:clustering}), setting both \mintinline{python}{min_samples=2} and \mintinline{python}{min_cluster_size=2} is tolerable. Applying these values on the \textbf{merged datasets} (5786 tweets) triggers \textbf{45,37\%} noise (2625 tweets).

\subsubsection{Iterative clustering}
\label{subsubsec:iterative_clustering}

\begin{figure}
    \centering
    \includegraphics[width=\textwidth]{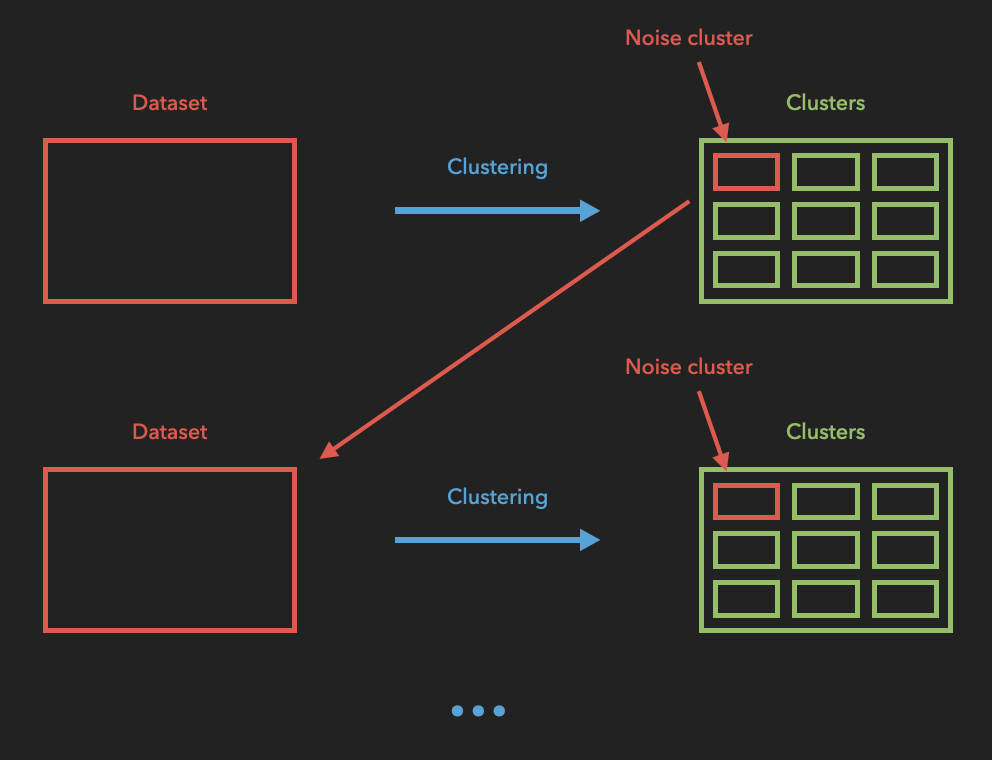}
    \caption{Scheme representing the iterative clustering principle.}
    \label{fig:iterative_clustering_scheme}
\end{figure}

Even when the parameters are adjusted to minimize the noise rate, it remains really high (cf \ref{subsubsec:parameters_selection}). Although knowing which time series have unique dynamics is important, such noise rate is excessive and may come from disproportionate proximity requirements. To overcome that, an \textit{iterative clustering} is proposed. Basically, each iteration consists in \textbf{considering the noise cluster as a dataset itself} on which the HDBSCAN algorithm is applied (figure \ref{fig:iterative_clustering_scheme}). At each round the density notion is redefined since the number of inputs is lower, so new clusters emerge. As they are of \textbf{poorer quality}, the round in which they appeared is stored to be reminded when clusters are shown.

The iteration stops when the noise share is bellow 5\% of the entire dataset.

\subsection{Results}
\label{subsec:first_results}

In \ref{appendix:clustering_results}, the 18 first clusters obtained from each dataset are displayed (figure \ref{fig:iterative_clustering_result_dataset_1}, \ref{fig:iterative_clustering_result_dataset_2} and \ref{fig:iterative_clustering_result_both_datasets}). Table \ref{tab:first_results} summarizes information about the clusters obtained. \\

\begin{table}[H]
    \centering
    \begin{tabular}{|m{25mm}|c|c|c|}
    \hline
        \textbf{Dataset} & \textbf{1} & \textbf{2} & \textbf{Both} \\ \hline
        \textbf{Number of tweets} & 2785 & 3001 & 5786 \\ \hline
        \textbf{Number of clusters} & 491 & 498 & 975 \\ \hline
        \textbf{Noise rate} & 1,1\% & 1,0\% & 1,4\% \\ \hline
        \textbf{Average cluster size} & 6 tweets & 6 tweets & 6 tweets \\ \hline
        \textbf{Standard deviation of the clusters size} & 9 tweets & 14 tweets & 17 tweets \\ \hline
        \textbf{Highest cluster size} & 141 tweets & 232 tweets & 349 tweets \\ \hline
    \end{tabular}
    \caption{Summary of the clustering results}
    \label{tab:first_results}
\end{table}

\begin{figure}
    \centering
    \includegraphics[width=\textwidth]{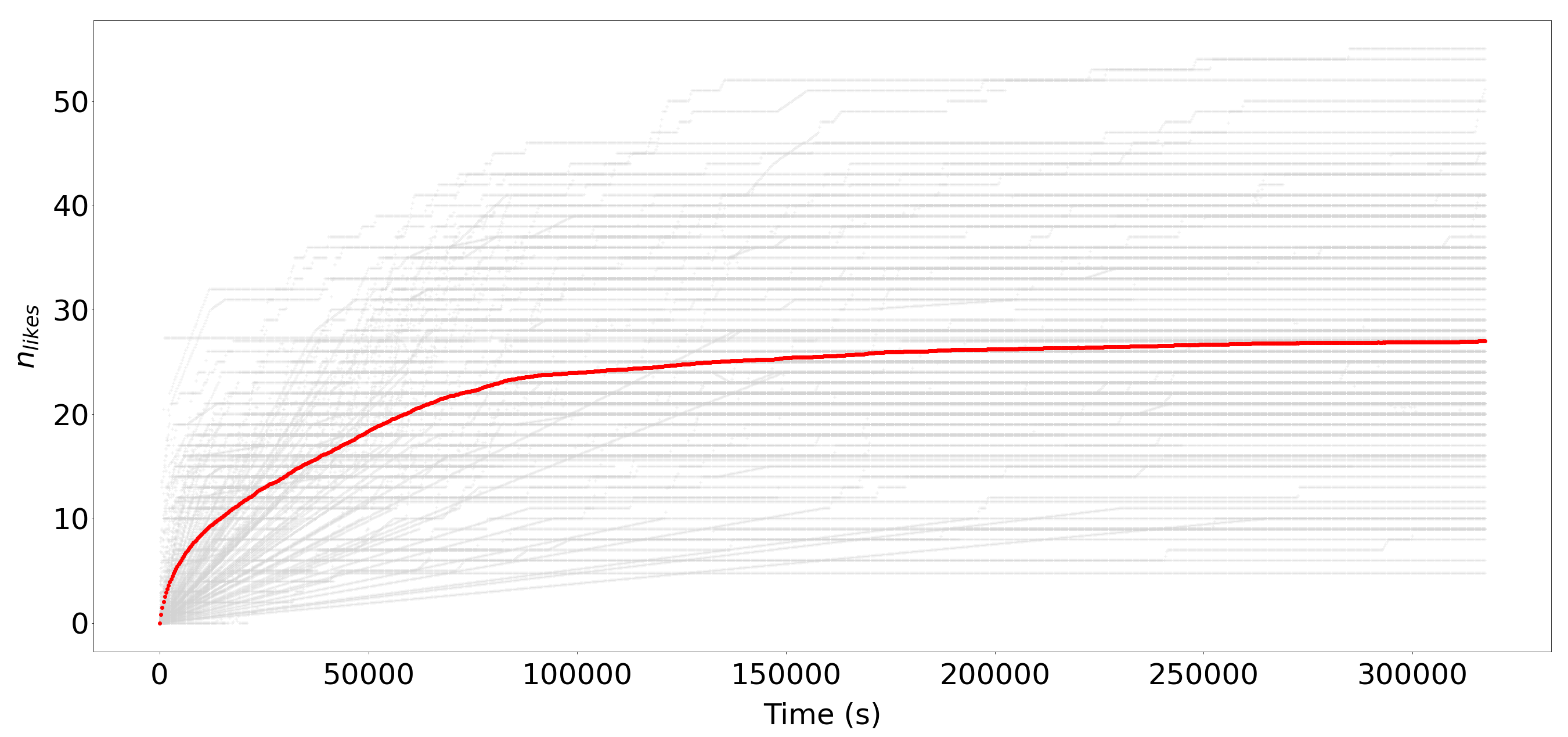}
    \caption{Biggest cluster from the $1^{st}$ dataset.}
    \label{fig:biggest_cluster_dataset4}
\end{figure}

Undoubtedly, these outcomes are not sufficient to answer our original problem (section \ref{subsec:problem}). The reasonable number of clusters is far from being reached and some of them are too disparate to be clearly describable (see figure \ref{fig:biggest_cluster_dataset4}). On top of that, clusters made of \textit{popular tweets} — those with high values of $n^{max}_{likes}$ — are often too small to be considered as pattern representatives, although this is to be expected given our parameters selection (section \ref{subsubsec:parameters_selection}). 

\noindent Several causes could explain such poor results :
\begin{itemize}
    \item the study might suffer from a \textbf{lack of data} : datasets might be too little for each pattern to be represented by an appropriate number of tweets ;
    \item discovering a reasonable amount of interpretable patterns \textbf{may not be possible} because of the plurality of social dynamics ;
    \item our \textbf{definition of evolution dynamics} itself (section \ref{sec:clustering}) may be too extreme or at least not accurate : so far, two curves have the \textit{exact same dynamics} if they are equal \textbf{at any point} (without considering the weighting).
\end{itemize}

In the remainder of this article, a \textbf{revision of this last point} is proposed, as it seems to be the most obvious source of error.

\section{THE TWEET VECTOR}
\label{sec:tweet_vector}

Instead of measuring how close popularity evolutions are at any point, the study will now focus on the \textbf{essential particularities} of what we call \textit{dynamics}. Then a clustering can be exclusively applied on those attributes. As it corresponds to a \textbf{relaxation} of our previous distance measure, it may lead to a lower amount of clusters without compromising their quality. \\

Every element that can characterize a time series' dynamics will be stored in a vector called the \textit{tweet vector}.

\subsection{Components of the tweet vector}
\label{subsec:vtweet_components}

The dynamics' characteristics identified are :
\begin{itemize}
    \item $n^{max}_{likes}$ — the maximum number of like — as it reflects the \textbf{intensity} of the dynamics ;
    \item $slope^{mean}$ — the average slope of the curve — since it describes \textbf{how fast} the dynamics are ;
    \item the \textit{key instants}, i.e. the times at which the evolution reaches a given percentage of its maximum value, to take into account its \textbf{temporal distribution} ;
    \item the \textit{boosts}' raw increases (see \ref{subsubsec:boosts}) which constitutes a \textbf{striking  distinction} between evolutions ;
    \item a compressed format of the L1 distance between the current curve and the others, which carries indications about its \textbf{overall (and relative) shape}.
\end{itemize}

Although $n^{max}_{likes}$ and $slope^{mean}$ are quite explicit values, the integration of the key instants, boosts and L1 distance components require some additional work. In doing so, we will attempt to obtain a tweet vector \textbf{as small as} possible, since the idea is to only carry necessary information.

\subsubsection{The key instants}
\label{subsubsec:key_instants}

It is assumed that extracting $t_p$ with $p \in \{10, 20, 30, 40, 50, 60, \\ 70, 80, 90\}$ (as manipulated is section \ref{fig:rudimentary_preprocessing}) is more than enough to apprehend the temporal distribution of the time series. The objective is to select the most relevant ones too reduce the tweet vector's size. To tackle this task, the well-known \textbf{Principal Component Analysis} (PCA) algorithm is used. Details about its theoretical functioning can be found in \cite{pca_wiki}.\\

The PCA's implementation of scikit-learn \cite{pca_lib} allows to be aware the share of variance kept during the dimension reduction process. It also offers the possibility to visualize a projection of the variance distribution of the different key instants over a given dataset. 

After running it on the dataset, it turns out that retaining the \textbf{3 first components allows to keep 95\% of the variance} and that the less correlated triplet of key instants is \textbf{($t_{10}$, $t_{50}$ and $t_{90}$)}. Although PCA's components are not equals to the key instants themselves, we can suppose this particular triplet contains the information we need.

\subsubsection{The boosts}
\label{subsubsec:boosts}

\begin{figure} [H]
    \centering
    \includegraphics[width=\textwidth]{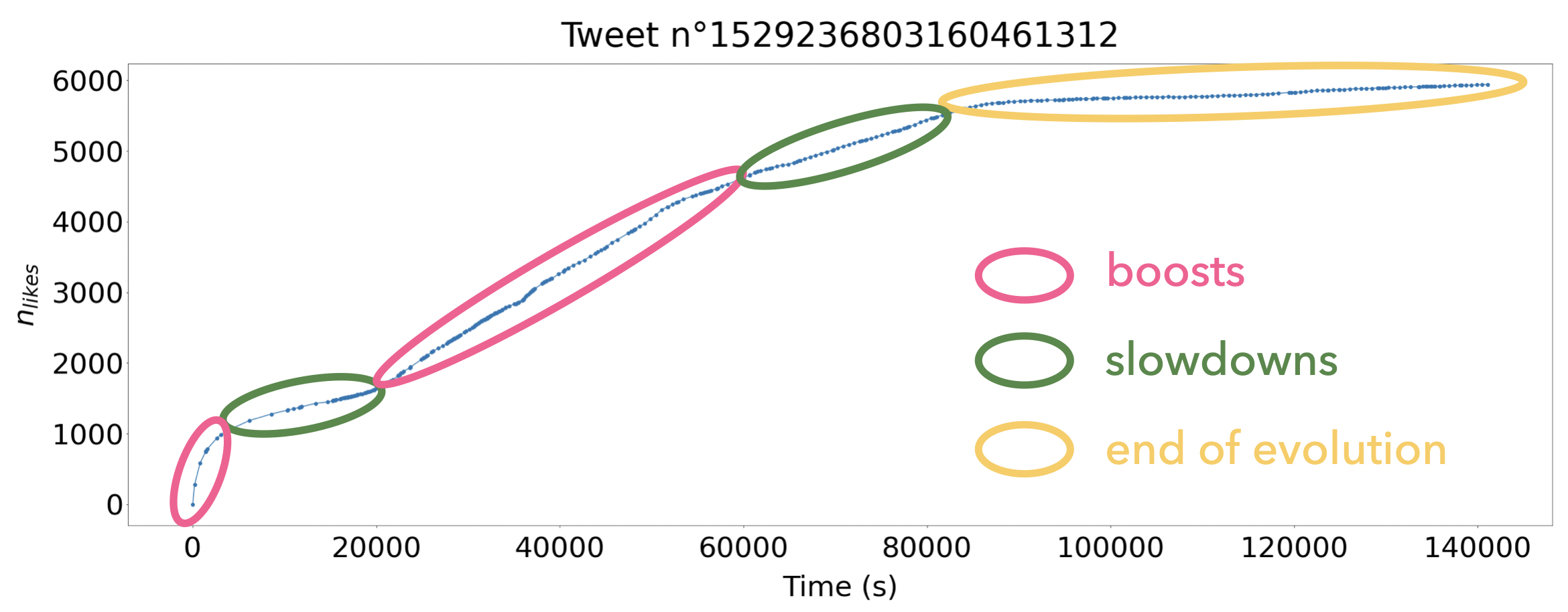}
    \caption{Qualitative identification of boosts in a given popularity evolution.}
    \label{fig:boost_definition}
\end{figure}

A \textit{boost} designates \textbf{a sudden change of rhythm favoring an increase of popularity}\footnote{boosts are inspired by the \textit{bursts} described in \cite{li_2016}} (see figure \ref{fig:boost_definition}). It is a specific behavior of the virality evolutions which could become an efficient tool to describe a given pattern.

\begin{figure} [H]
    \centering
    \includegraphics[width=\textwidth]{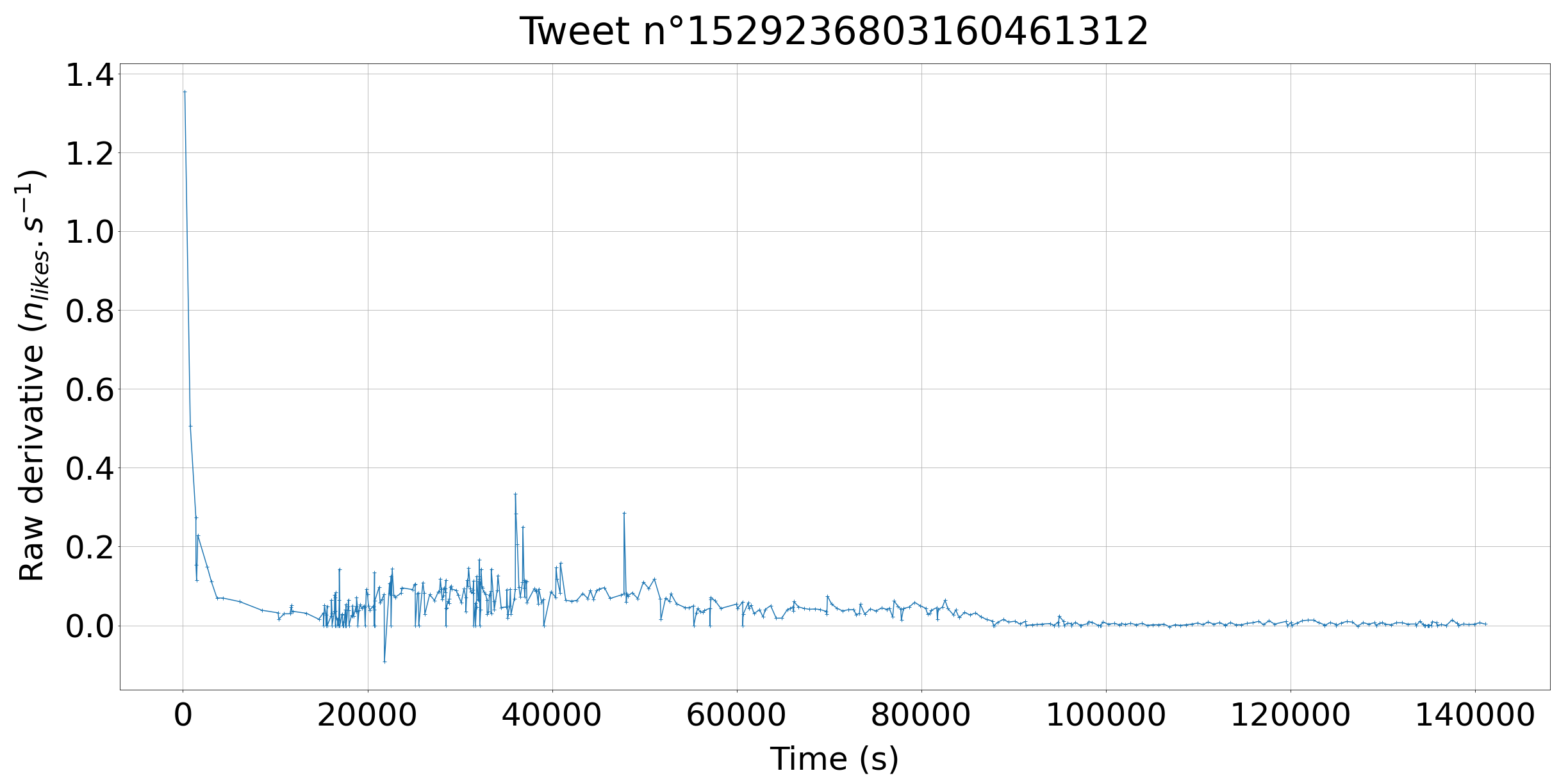}
    \caption{Raw derivative of the time series window : the second boost does not stand out enough to be identified.}
    \label{fig:raw_derivative}
\end{figure}

Instinctively, the mathematical definition must integrate a condition concerning the \textbf{derivative} of the evolution. However, data imperfections can make the task complicated (figure \ref{fig:raw_derivative}).

A consistent way to highlight boosts while smoothing the derivative is to employ \textbf{Total Variation Denoising (TV Denoising)}. This technique consists in approaching the derivative with a \textbf{piecewise constant curve}. This is equivalent to solving the following problem :

\begin{equation*}
    \text{find } \mathcal{R}_{min} = argmin\left(f(\mathcal{O}, .)\right) \\
\end{equation*}
\begin{equation*}
\text{where } f(\mathcal{O}, \mathcal{R}) = \norm{ \mathcal{O}-\mathcal{R} }_2 +         \lambda.\norm{ \nabla\mathcal{R} }_1
\end{equation*}

\noindent Here, $\mathcal{O}$ is the original derivative values, $\mathcal{R}$ is the vector representing the reconstructed derivative and $\lambda$ is a scalar affecting the "penalty" applied to the variation $\nabla\mathcal{R}$ : the higher $\lambda$ is, the less variations the reconstructed curve contains. In practice, this parameter is chosen so that \textbf{the maximum number of boosts in the dataset doesn't exceed 6}. Beyond 6, some boosts are likely to be mistakes caused by an excessive sensitivity. \\

Thanks to the \mintinline{python}{cvxpy} python library \cite{cvxpy_lib}, this method is easily implementable. \\

\begin{figure*}
    \centering
    \includegraphics[width=0.9\textwidth]{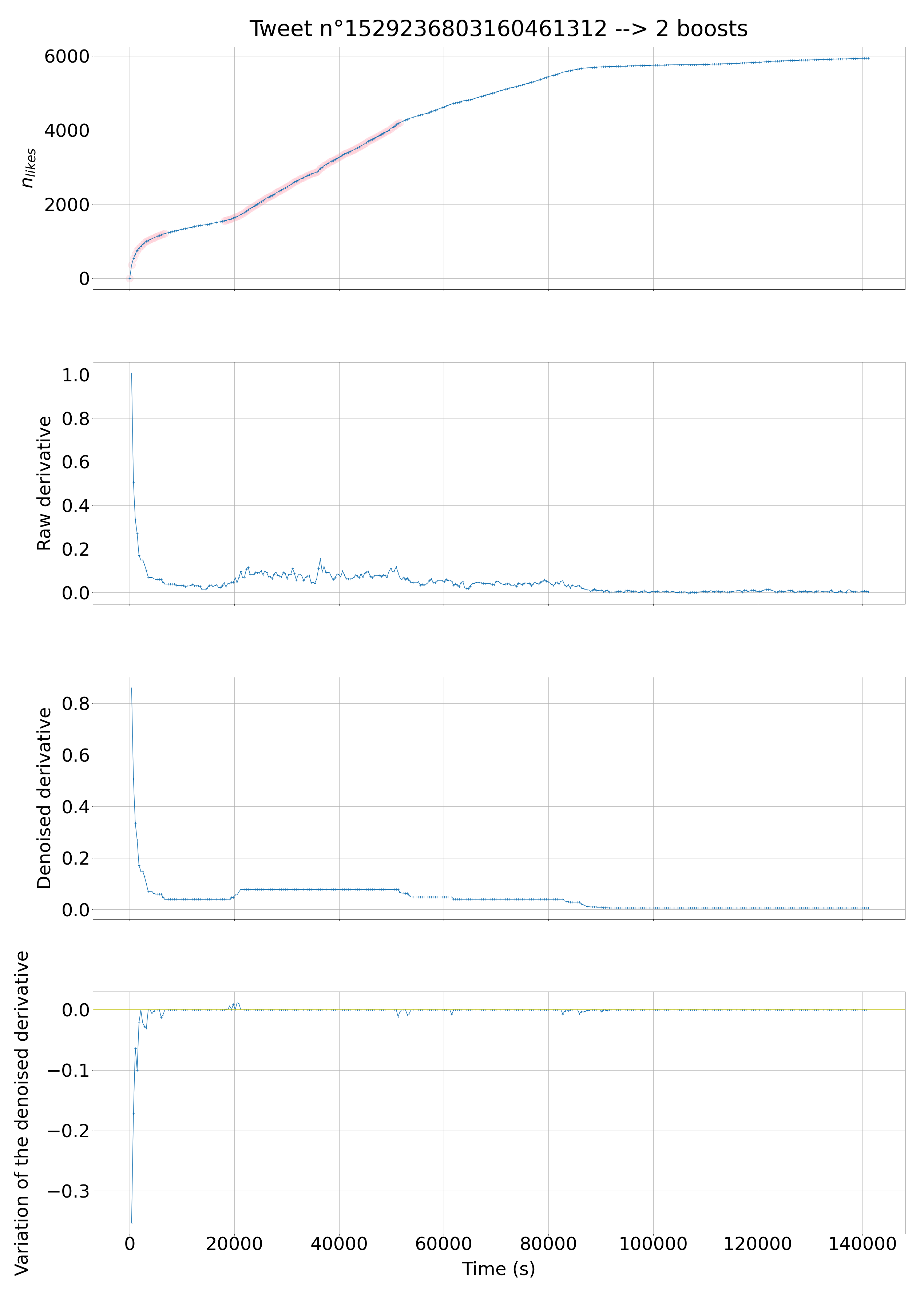}
    \caption{Encoding and identification of the boosts ($\lambda=0,6$).}
    \label{fig:boosts_identification}
\end{figure*}

Boosts correspond to the \textbf{denoised derivative's bumps}. In order to detect them properly, an \textbf{encoding} program is performed. First, the variations of the denoised derivative are computed. After removing the numerical noise, the signs of it are extracted. Hence, each tweet is related to a vector made of 1, -1 and 0. Finally, the boosts are enumerated and identified knowing that they start with a 1 and end as soon as a -1 is encountered. Moreover, if the first (resp. the last) sign is -1 (resp. 1), it means the curve begins (resp. ends) with a boost. Figure \ref{fig:boosts_identification} illustrates the different steps of the process and spotlights the resulting boosts. \\

The raw increase of boosts is added to the tweet vector. Hence, 6 components should be dedicated to it with the $i^{th}$ component representing the $i^{th}$ boost of the evolution (or equal to 0 if there is no $i^{th}$ boost). However, the \textbf{3 first boosts seem to be the most determining} : a tweet having 4 boosts is often almost inert, so that a single like is considered as a great increase. That is why only \textbf{three components are eventually allocated to this characteristic}.

\subsubsection{The L1 distance integration}
\label{subsubsec:L1_distance_components}

\begin{figure} [H]
    \centering
    \includegraphics[width=\textwidth]{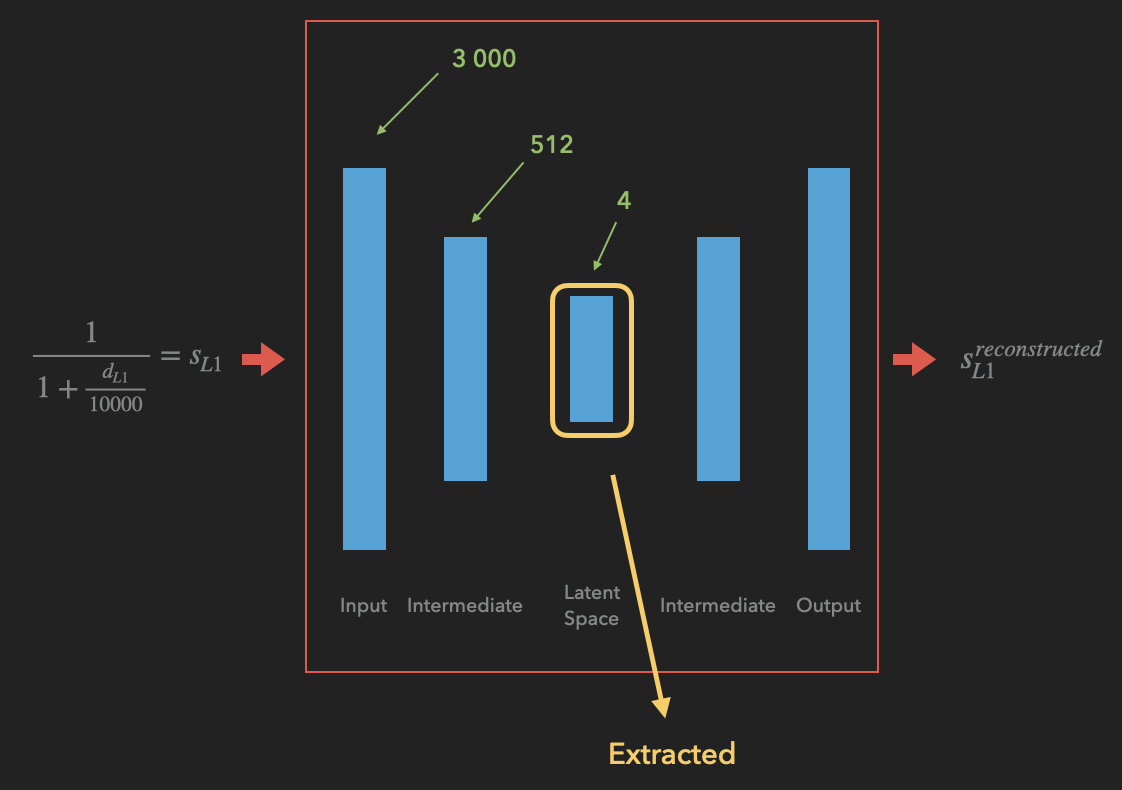}
    \caption{Scheme representing the model used on dataset 2. It consists of a 512-neurons dense hidden layer and must be fed with a similarity to function optimally.}
    \label{fig:auto_encoder}
\end{figure}

To be integrated to the tweet vector, the L1 distance matrix — containing all the pairwise L1 distances of the dataset — is reduced with an \textbf{auto-encoder neural network}. Without detailing the fundamentals of deep learning, this model is built with two symmetrical networks : the \textit{encoder} and the \textit{decoder}. The first one is trained to encrypt the data in a lower dimensional space while the second one is trained to reconstruct the initial data based on the encryption (also called \textit{latent space}). If the decoder succeeds in retrieving the original data, it means the latent space \textbf{carries the essential information} of the matrix. The architecture of our model for the $2^{nd}$ dataset is shown in figure \ref{fig:auto_encoder} : for each tweet, \textbf{4 components} are enough to describe its distance to all the others. The ELU activation function \cite{elu_pytorch} and smooth L1 loss function \cite{smoothl1loss_pytorch} are used.

\subsection{Clustering on the tweet vector}
\label{subsec:vtweet_clustering}

Once all the attributes related to popularity dynamics are gathered, the tweet vector $\Pi$ have the following form :

\begin{equation*}
    \Pi = (n_{likes}^{max}, slope_{mean}, t_{10}, t_{50}, t_{90}, n^1_{boost}, n^2_{boost}, n^3_{boost}, d_1, d_2, d_3, d_4)
\end{equation*}

\noindent where $n^i_{boost}$ is the raw increase of the $i^{th}$ boost and $d_j$ is the $j^{th}$ component of the reduced L1 distance matrix for the current tweet. \\

Now, the HDBSCAN algorithm is applied to the tweet vector — the clustering distance being the \textbf{euclidean distance} between the components — in order to obtain better results than before (section \ref{subsec:first_results}). The clustering parameters \mintinline{python}{min_cluster_size} and \mintinline{python}{min_sample} are set to the same values as previously (i.e. 2) to allow comparison between the two methods.

\subsection{Results}
\label{subsec:vtweet_results}

\begin{table}[H]
    \centering
    \begin{tabular}{|m{25mm}|c|c|c|}
    \hline
        \textbf{Dataset} & \textbf{1} & \textbf{2} & \textbf{Both} \\ \hline
        \textbf{Number of tweets} & 2785 & 3001 & 5786 \\ \hline
        \textbf{Number of clusters} & 341 & 329 & 690 \\ \hline
        \textbf{Noise rate} & 43,7\% & 41,0\% & 42,6\% \\ \hline
        \textbf{Average cluster size} & 8 tweets & 9 tweets & 8 tweets \\ \hline
        \textbf{Standard deviation of the clusters size} & 65 tweets & 68 tweets & 93 tweets \\ \hline
    \end{tabular}
    \caption{Summary of the tweet vector clustering results}
    \label{tab:vtweet_results}
\end{table}

The 12 first clusters for both dataset 1 and 2 are displayed in \ref{appendix:vtweet_results}. \\

Clearly, the use of the tweet vector hasn't improved the clustering significantly. Worse, it has probably even degraded it. Indeed, the number of clusters has not reduced — the value exposed here is lower but it must be kept in mind that the iterative clustering has not been applied — and their composition is sometimes surprising (see the $7^{th}$ cluster). \\

Nevertheless, the boost detection tool remains helpful to characterize the different evolutions.

\section{OVERALL ANSWERS AND AVENUES FOR REFLEXION}
\label{sec:retrospection}

In the end, this study \textbf{have not allowed us} to find a reasonable number of easily interpretable patterns in tweets popularity evolution. The naive approach, which interpreted the dynamics closeness as a point to point distance, enabled to identify the main difficulties while providing an initial overview of the problem. Thanks to this fast and simple construction, the importance of \textbf{accurately defining the elements manipulated} and \textbf{properly quantifying the expected results} has been understood. The revision that followed tried to satisfy those new requirements through the use of a \textit{tweet vector}, whose components are said to characterize the popularity dynamics. Unfortunately, this sort of "feature extraction" didn't lead to encouraging results. \\

However, this strategy is far from being fully explored. On the one hand, the effect of \mintinline{python}{min_cluster_size} and \newline \mintinline{python}{min_samples} has been underestimated. A solid method to choose them unambiguously would be of great interest since they exercise strong control over the amount of clusters. On the other hand, the tweet vector features are likely to be inappropriate : a more elaborate work focused on the boosts or similar objects may bring promising outcomes. \\

Whether or not a positive response can be provided to the initial problem therefore depends essentially on the ability to identify the right features on which the clustering algorithm is applied.


\printbibliography

\newpage

\appendix

\section{Keywords and users associated to datasets}
\label{appendix:datasets}

Dataset 1 : \\

\begin{tabular}{|c|c|c|}
    \hline
    \textbf{Keywords} & \textbf{Hashtags} & \textbf{Users} \\ \hline
    \textit{Ukraine} & \textit{\#TopGun} & \textit{@EmmanuelMacron} \\ \hline
    \textit{France} & & \textit{@elonmusk} \\ \hline
    \textit{Foot} & & \textit{@Thom\_astro} \\ \hline
\end{tabular} \\

\vspace{1cm}

\noindent Dataset 2 : \\

\begin{tabular}{|c|c|}
    \hline
    \textbf{Hashtags} & \textbf{Users} \\ \hline
    \textit{\#canicule} & \textit{@fetemusique} \\ \hline
    \textit{\#Ukraine} & \textit{@NUPES\_2022\_} \\ \hline
    \textit{\#Sievierodonetsk} & \textit{@top14rugby} \\ \hline
    \textit{\#Luhansk} & \textit{@KyivIndependent} \\ \hline
    \textit{\#legislatives2022} & \textit{@ActuFoot\_} \\ \hline
    \textit{\#Législatives} & \textit{@AP} \\ \hline
    \textit{\#F1} & \textit{@elonmusk} \\ \hline
    \textit{\#CanGP} & \textit{@AllanBARTE} \\ \hline
    \textit{\#CanadaGP} & \textit{@netflix} \\ \hline
    \textit{\#FormulaOne} & \textit{@NASA} \\ \hline
    \textit{\#MontrealGP} & \\ \hline
    \textit{\#Taiwan} & \\ \hline
    \textit{\#FathersDay} & \\ \hline
    \textit{\#FeteDeLaMusique} & \\ \hline
    \textit{\#NUPES} & \\ \hline
    \textit{\#LREM} & \\ \hline
    \textit{\#Climate} & \\ \hline
    \textit{\#ClimateEmergency} & \\ \hline
    \textit{\#ClimateChange} & \\ \hline
    \textit{\#HumanRights} & \\ \hline
    \textit{\#RefugeesDay} & \\ \hline
    \textit{\#Glastonbury2022} & \\ \hline
    \textit{\#glastonburyfestival} & \\ \hline
    \textit{\#Bitcoin} & \\ \hline
    \textit{\#EndGunViolence} & \\ \hline
    \textit{\#KevinSpacey} & \\ \hline
    \textit{\#COVID19} &  \\ \hline
\end{tabular}

\section{Clustering results with the naive approach}
\label{appendix:clustering_results}

\begin{figure*}
    \centering
    \includegraphics[width=\textwidth]{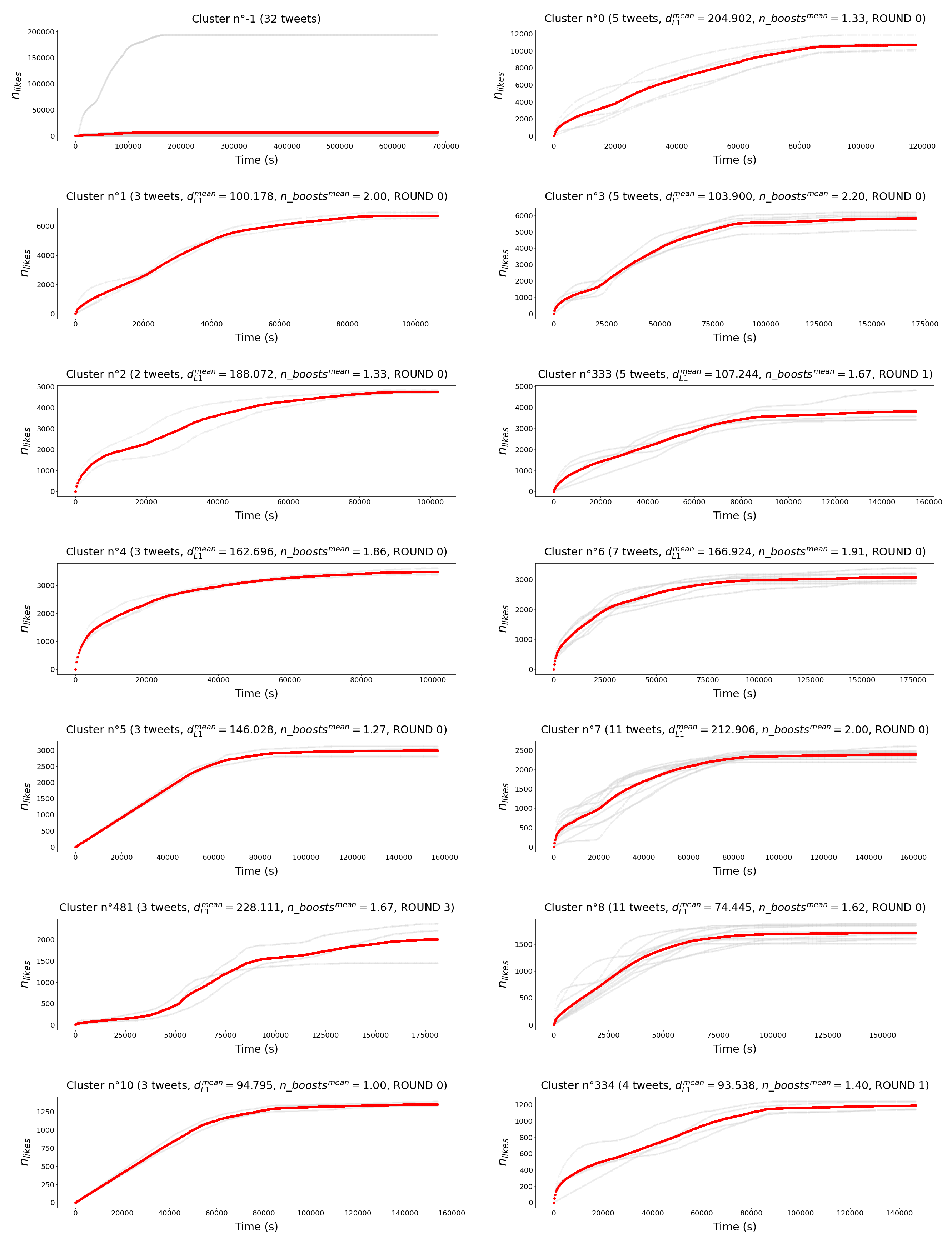}
    \caption{The first 18 clusters (sorted by $n^{max}_{likes}$) obtained from dataset 1 : $d^{moy}_{L1}$ represents the average L1 distance within each cluster, $n\_boosts^{mean}$ stands for the average amount of boosts (cf section \ref{subsubsec:boosts}) in each cluster and ROUND symbolizes the iteration number during which the cluster appeared. The red line is the average of all the gray curves (which are the actual time series).}
    \label{fig:iterative_clustering_result_dataset_1}
\end{figure*}

\begin{figure*}
    \centering
    \includegraphics[width=\textwidth]{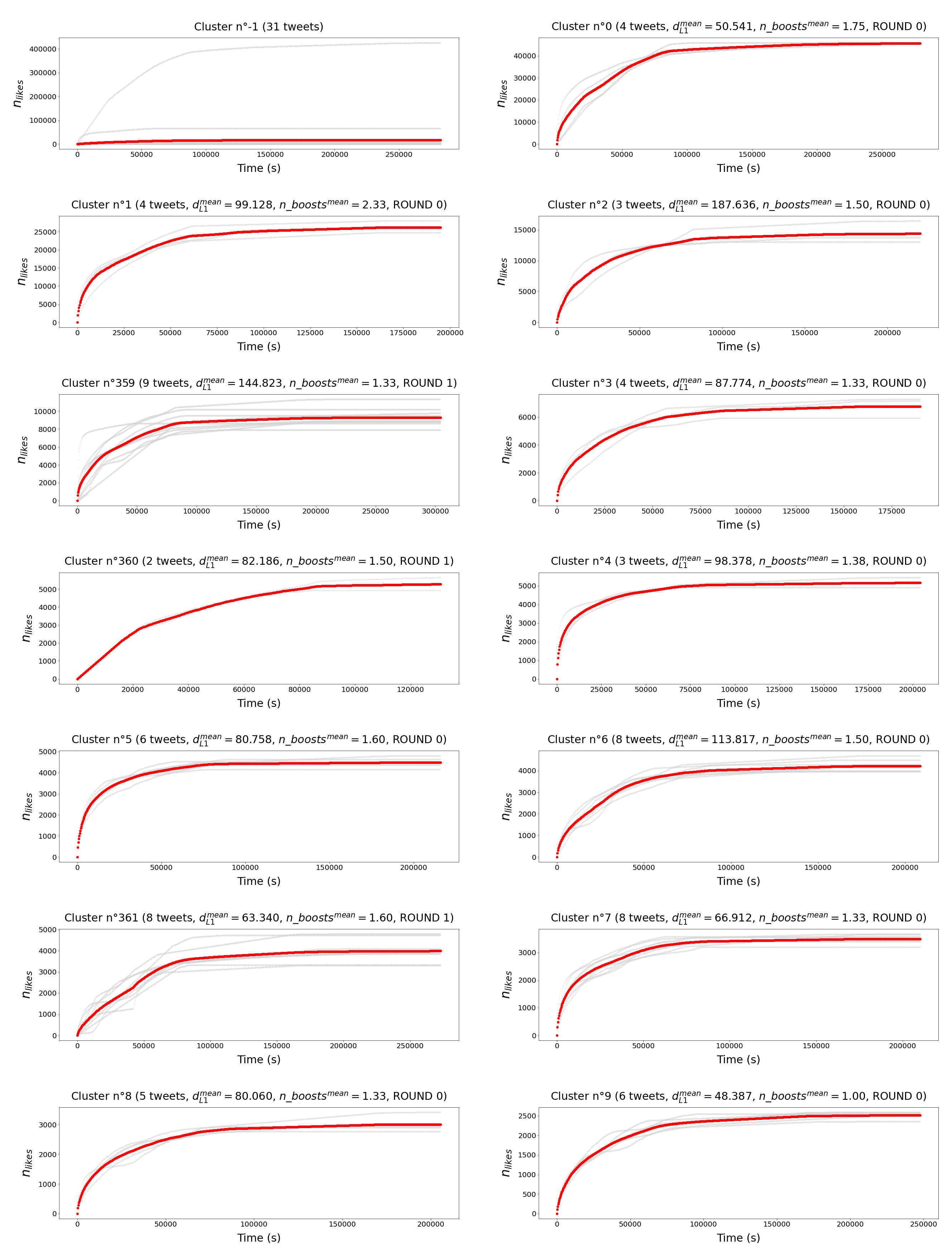}
    \caption{The first 18 clusters (sorted by $n^{max}_{likes}$) obtained from dataset 2 : $d^{moy}_{L1}$ represents the average L1 distance within each cluster, $n\_boosts^{mean}$ stands for the average amount of boosts (cf section \ref{subsubsec:boosts}) in each cluster and ROUND symbolizes the iteration number during which the cluster appeared. The red line is the average of all the gray curves (which are the actual time series).}
    \label{fig:iterative_clustering_result_dataset_2}
\end{figure*}

\begin{figure*}
    \centering
    \includegraphics[width=\textwidth]{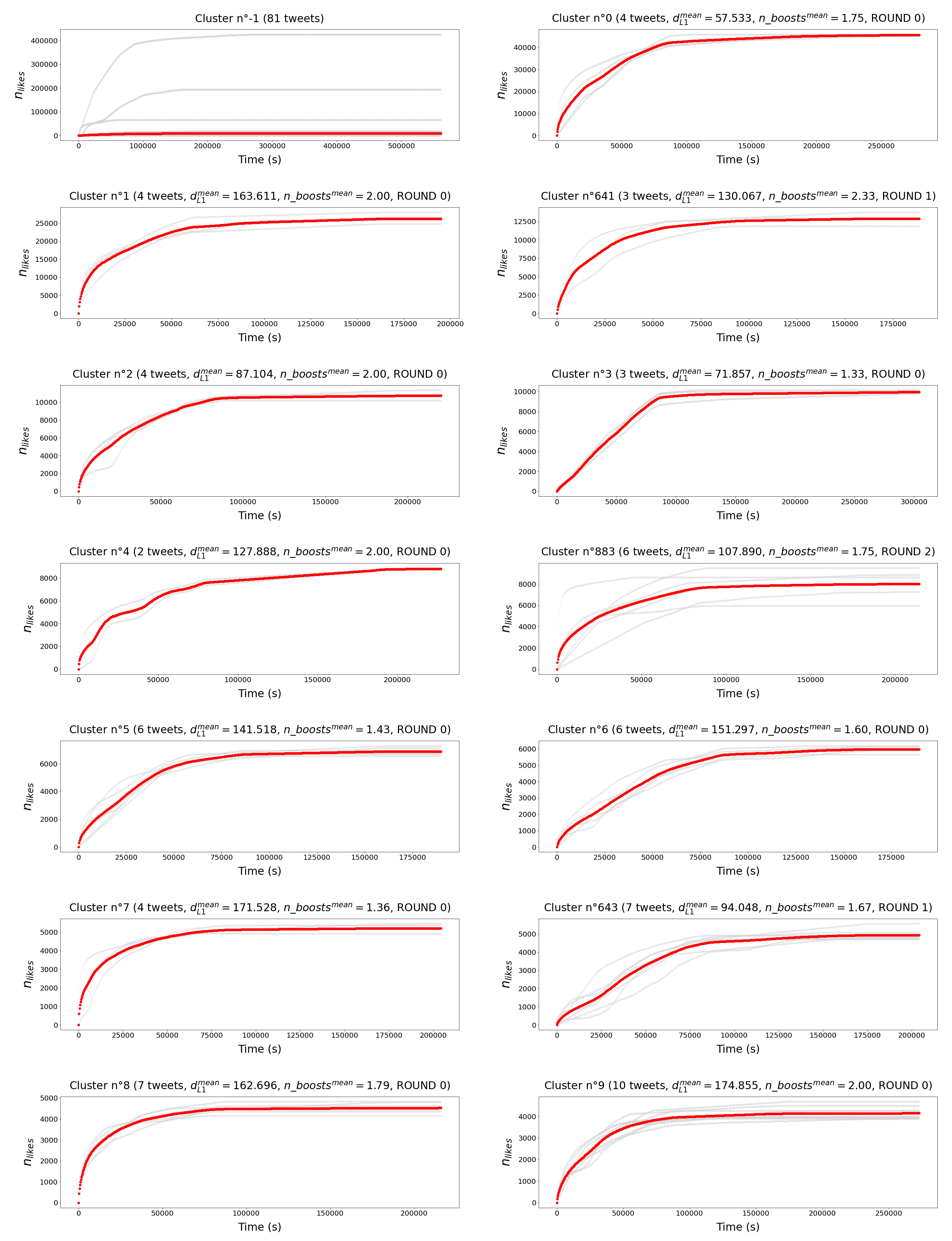}
    \caption{The first 18 clusters (sorted by $n^{max}_{likes}$) obtained from both dataset 1 and 2 : $d^{moy}_{L1}$ represents the average L1 distance within each cluster, $n\_boosts^{mean}$ stands for the average amount of boosts (cf section \ref{subsubsec:boosts}) in each cluster and ROUND symbolizes the iteration number during which the cluster appeared. The red line is the average of all the gray curves (which are the actual time series).}
    \label{fig:iterative_clustering_result_both_datasets}
\end{figure*}

\section{Tweet vector clustering results}
\label{appendix:vtweet_results}

\begin{figure*}
    \centering
    \includegraphics[width=\textwidth]{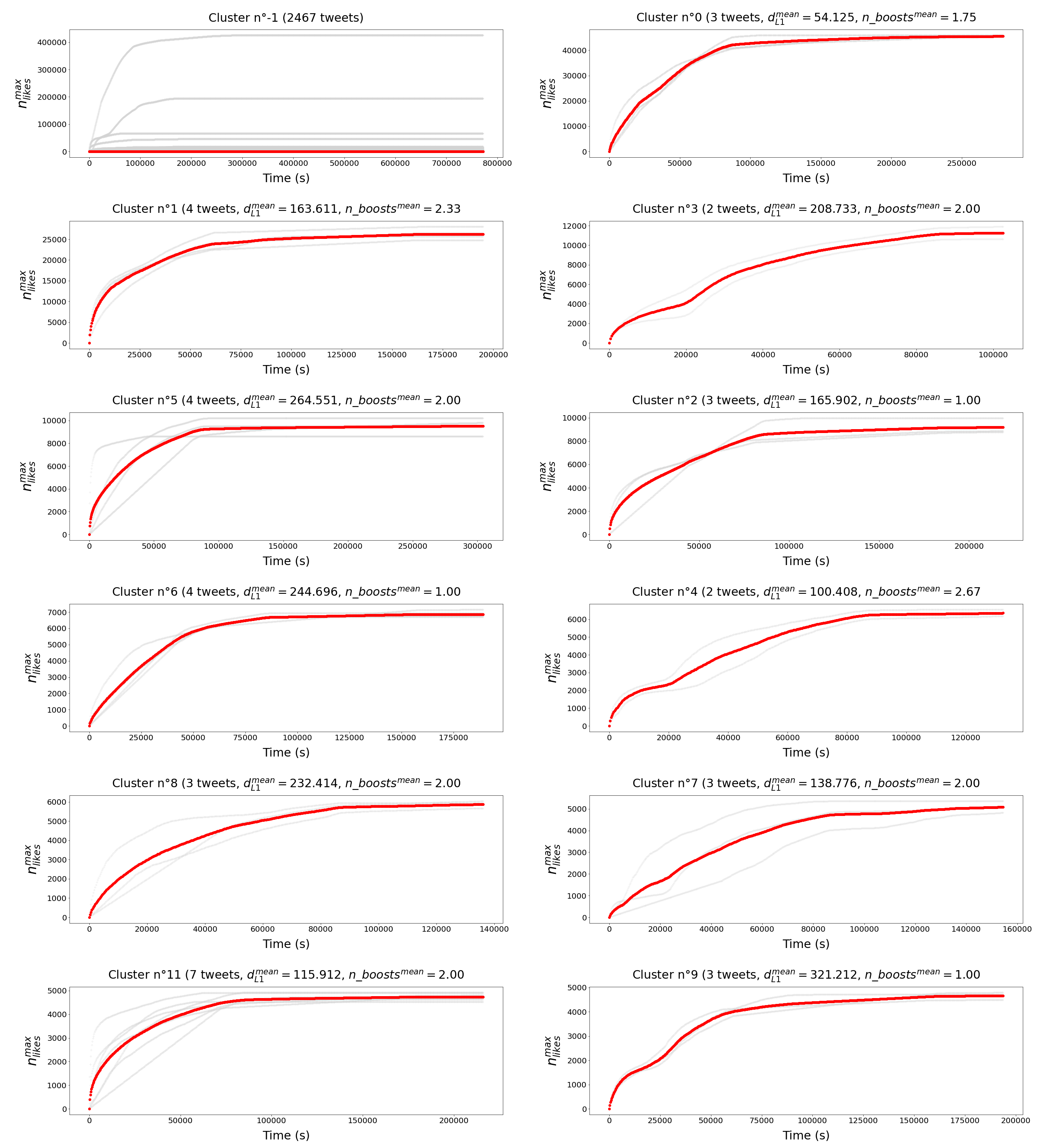}
    \caption{The first 12 clusters (sorted by $n^{max}_{likes}$) obtained from the \textbf{tweet vectors} of both dataset 1 and 2.}
    \label{fig:vtweet_clustering_result_both_datasets}
\end{figure*}

\end{document}